\documentclass[lettersize,journal]{IEEEtran}

\usepackage[pagebackref=false,breaklinks=true,linkcolor=red,anchorcolor=blue, citecolor=green,colorlinks,bookmarks=false]{hyperref}
\usepackage{amsmath,amsfonts}
\usepackage{algorithmic}
\usepackage{array}
\usepackage[caption=false,font=normalsize,labelfont=sf,textfont=sf]{subfig}
\usepackage{textcomp}
\usepackage{stfloats}
\usepackage{url}
\usepackage{verbatim}
\usepackage{graphicx}
\usepackage{tabularx}
\usepackage{mathtools}
\usepackage{xcolor}

\def\BibTeX{{\rm B\kern-.05em{\sc i\kern-.025em b}\kern-.08em
    T\kern-.1667em\lower.7ex\hbox{E}\kern-.125emX}}
\usepackage{balance}

\newcommand{\ie}{{\it i.e. }}
\newcommand{\fl}{\mathcal{L}}
\newcommand{\emean}{\mathbb{E}}

\begin{document}
\title{An End-to-End Depth-Based Pipeline for Selfie Image Rectification}
\author{Ahmed Alhawwary, Janne Mustaniemi, Phong Nguyen-Ha, Janne Heikkilä
\thanks{A. Alhawwary, J. Mustaniemi, and J. Heikkilä are with the Center for Machine Vision and Signal Analysis
(CMVS), University of Oulu, 90570 Oulu, Finland.}
\thanks{P. Nguyen-Ha is with Qualcomm AI Research, 10000 Ha Noi, North, Vietnam.}
}

\maketitle

\begin{abstract}

Portraits or selfie images taken from a close distance typically suffer from perspective distortion. 
In this paper, we propose an end-to-end deep learning-based rectification pipeline to mitigate the effects of perspective distortion. We learn to predict the facial depth by training a deep CNN. The estimated depth is utilized to adjust the camera-to-subject distance by moving the camera farther, increasing the camera focal length, and reprojecting the 3D image features to the new perspective. The reprojected features are then fed to an inpainting module to fill in the missing pixels. We leverage a differentiable renderer to enable end-to-end training of our depth estimation and feature extraction nets to improve the rectified outputs.  To boost the results of the inpainting module, we incorporate an auxiliary module to predict the horizontal movement of the camera which decreases the area that requires hallucination of challenging face parts such as ears. Unlike previous works, we process the full-frame input image at once without cropping the subject's face and processing it separately from the rest of the body, eliminating the need for complex post-processing steps to attach the face back to the subject's body. To train our network, we utilize the popular game engine Unreal Engine to generate a large synthetic face dataset containing various subjects, head poses, expressions, eyewear, clothes, and lighting. Quantitative and qualitative results show that our rectification pipeline outperforms previous methods, and produces comparable results with a time-consuming 3D GAN-based method while being more than 260 times faster.
\end{abstract}


\section{Introduction}
\IEEEPARstart{N}{owadays,} mobile phones are equipped with high-quality and advanced cameras which are vastly popular for capturing selfies or close-range portraits. Nevertheless, the images shot from a close distance suffer from perspective distortion. This problem happens regardless of the camera specification, quality or lens, but due to the inherent nature of the perspective projection when the object is relatively close to the camera. While the camera type is not the reason behind this kind of distortion, it is tied to the wide-angle cameras because they enjoy a wide field of view to fit the face (and possibly parts of the body) from a close distance (around 20-80 cm) which is the typical case for hand-captured selfies. In facial perspective distortion, the face parts are disproportionate; the nose is popping out and looks bigger and the face looks squashed backwards as shown in Figure \ref{fig:X_to_5X}. In addition to the possibly unpleasant visual appearance, prior research \cite{zhao2019learning,kao2023toward} revealed that applications such as face verification and reconstruction are highly affected by perspective distortion.

Since the term face distortion is used ambiguously in the related literature, it is worth noting that the perspective distortion we are addressing here is different from the distortion caused by the camera lens or optics such as radial distortion which bends shapes and straight lines on image boundaries. Perspective projection also causes stretches to faces and subjects when they are close to image edges. This becomes more prominent when the field-of-view is more than 72° \cite{lai2021correcting, shih2019distortion} and occurs due to the wide angle of the projection. While our method can deal with perspective distortion in general,  the techniques used to correct stretch distortion can not handle close-range perspective distortion as it has different characteristics. More specifically, those approaches aim to fix the face stretches when being near the image boundaries but not the other distortion characteristics resulting from the close distance to the camera.

While various types of facial distortion have been explored in literature, there has been limited focus on perspective distortion in close-range portraits. In this paper, we propose a depth-based solution to rectify the perspective distortion. We learn to estimate the facial depth of the distorted image by training a deep convolutional neural network (CNN). The estimated depth map is used to un-project the extracted image features to the 3D coordinates of the camera.  We then adjust the camera-to-subject distance by virtually moving the camera farther from the 3D subject while increasing its focal length proportionally to maintain the same scale and reproject it to the new camera view. The holes and missing pixels resulting from warping are filled by a subsequent inpainting module.  We train the feature extractor and the depth module in an end-to-end (E2E) fashion by utilizing a differentiable rendering module \cite{ravi2020pytorch3d}, and apply the warping in feature space instead of mere RGB space.

 Prior methods either rely on facial landmarks for reconstruction \cite{fried2016perspective}, which has been shown to be error-prone \cite{zhao2019learning, kao2023toward}, or train a CNN  separately to predict a 2-channel deformation map \cite{zhao2019learning}.  Recently, Wang et al. proposed the Disco method \cite{wang2023disco} leveraging the recent image generation capabilities of pretrained 3D Generative Adversarial Networks (GANs). However, it is time-consuming and requires several minutes on an advanced GPU for the inversion process which estimates the latent face code and camera parameters corresponding to a crop of the distorted input image. Our experiments show that our method outperforms the approaches of Freid \cite{fried2016perspective} and LPUP \cite{zhao2019learning}  both quantitatively and qualitatively, and produces comparable results with Disco \cite{wang2023disco} while being more than 260 times faster.

Unlike previous methods, our solution works without cropping the input image first eliminating the necessity for complicated post-processing steps to combine the face back to the body as in Disco \cite{wang2023disco}. LPUP \cite{zhao2019learning} does not even address the problem of combining the face back to the rest of the body.
 
To boost the quality of the image produced by the generator (inpainting) module, we further incorporate an auxiliary module to predict the horizontal translation of the camera that decreases the area that needs to be filled.  This means that the camera might be translated horizontally along the X-axis in addition to the translation along the Z-axis (\ie the optical axis) as shown by the examples illustrated in Figure \ref{fig:horiz_missing}.  

To train our pipeline, we need a large amount of perspective-distorted and distortion-free image pairs. LPUP \cite{yin2021learning} mainly utilizes an online-available dataset BU-4DFE \cite{zhang2013high} consisting of RGBD facial images with limited head orientation, fixed expressions and illumination, and the distorted images are rendered simply using perspective projection and rasterization. Alternatively, we utilize the game development software Unreal Engine  \cite{unrealengine}, which includes high-quality characters known as Metahuman, to synthesize photorealistic pairs of close and far-range images. During data generation, we vary the expressions, grooms, eyeglasses, head orientation, camera-head poses, illumination and outfits. Our experiments demonstrate the generalization capabilities of our pipeline trained with synthetic data to handle the distortion in real-world images.

Our contributions are summarized as follows: 
\begin{itemize}
    \item  An E2E depth-based perspective rectification pipeline that outperforms previous solutions, and produces comparable results with a recent time-consuming, 3D GAN-based method Disco \cite{wang2023disco} while being more than 260 times faster. 
    \item  An approach for estimating the horizontal camera movement that reduces the region of missing pixels to be predicted.
    \item  A synthetic dataset of close and far-range image pairs created using Unreal Engine for training and testing our rectification pipeline. 
    % \item We will release the source code, trained models, and the dataset upon the publication of the paper.   

\end{itemize}

\section{Related Work}

Our work is deemed part of a wider area of a computer vision task known as novel view synthesis (NVS), where a novel view with a new perspective is computed from single or multiple reference images. Moreover, our work here is closely related to a sub-task within NVS called depth image-based rendering (DIBR) \cite{luo2016hole,shih20203d,wiles2020synsin}. In DIBR, an image is rendered from a new perspective leveraging the scene depth. In this work, we learn to estimate the facial depth of perspective-distorted images by training a deep CNN.  The depth is used to reproject the subject to a novel camera view to rectify the distortion.

The prior works that directly address the perspective distortion of close-range portrait can be categorized into 3 classes: Conventional optimization-based \cite{fried2016perspective}, learning-based \cite{zhao2019learning, nagano2019deep}, and 3D GAN-based methods \cite{wang2023disco}. Fried \cite{fried2016perspective} adjusts the distance between the camera and subject given a single portrait by estimating an initial head model and camera intrinsic and extrinsic parameters where they
are jointly optimized to find the optimal head model, expression and camera
parameters that fit pre-detected facial landmarks. This 3D model is then used
to compute a 2D smooth, pixel-wise motion field which produces no holes after warping the original image. 

However, in LPUP \cite{zhao2019learning}, they argued that utilizing 2D landmarks in
perspective-distorted images is suboptimal. Alternatively, they proposed a deep learning-based approach, where they train a network to predict a depth label corresponding to a specific depth range. This label is then provided to a flow estimation
network along with the face crop of the input image to estimate a 2D pixel-motion map that
corrects the distortion in the image. The warping produces holes that are filled by
a subsequent separate inpainting network. All their subnetworks are independently
trained and cascaded. On the contrary, our depth estimation network is trained in an E2E fashion by utilizing a differentiable projection module \cite{ravi2020pytorch3d}. Moreover, our pipeline estimates a depth map for the foreground subject that is used to compute the new pixel positions (\ie the deformation field) after changing the camera pose instead of training the network to predict the motion field directly. Estimating a 2-channel warping field introduces an unnecessary degree of freedom for the network while the motion can be derived from the depth map. Besides, our depth-based formulation enables us to freely control the camera displacement by changing the camera's extrinsic parameters. This allows us to manipulate the horizontal camera translation and move the camera in the direction that reduces the missing region as illustrated in Figure \ref{fig:horiz_missing}, which is not achievable using LPUP \cite{zhao2019learning}.  Additionally, LPUP requires the face to be cropped and scaled in the same way as training data, and they do not address the problem of combining the cropped face back to the body. However, in our approach, we deal with the whole subject eliminating the need to postprocess the head with the rest
of the body afterwards. Another difference is that our warping is performed on the extracted image features instead of the mere RGB image space thanks to the differentiable projection module \cite{wiles2020synsin, ravi2020pytorch3d}.

Regarding the training dataset, LPUP \cite{yin2021learning} synthesizes training samples from a static 3D head model captured by a light-stage system. However, this model has fixed expressions, head orientation and limited representation of the subject's hair. To increase the number of samples, they utilize a publicly available dataset of faces with their corresponding depth maps \cite{zhang2013high}. This dataset also has limited head orientation and mainly captures the subject looking forward to the camera along with fixed expression and illumination.  The distorted images were then rendered simply from the RGBD images using perspective projection and rasterization, without considering changes in illumination or complex reflectance properties \cite{yin2021learning}. In contrast, we leverage the popular game engine Unreal Engine (UE) to generate photorealistic image pairs with various expressions, eyeglasses, grooms, head orientation, camera-head poses, illumination and body clothes. 

Deep Face Normalization (DFN) \cite{nagano2019deep} proposed a module to rectify the perspective distortion as part of a pipeline aimed at 'normalizing' facial appearances. While their module shares similarities with LPUP \cite{zhao2019learning} in that it also generates 2D motion maps, DFN adopts a simpler approach. Specifically, it consists of a single network that takes the distorted face image along with its 2D facial landmarks as input.

Recently, Disco \cite{wang2023disco} was proposed to adjust the
camera-to-subject distance with the help of 3D GANs. They rely on the recent powerful generation capability of a pretrained 3D GAN, known as  EG3D \cite{chan2022efficient}, to generate plausible face corrections.  The 3D GANs are a type of generative network that is
additionally conditioned on camera parameters to generate appearance-consistent facial images under different view angles \cite{chan2022efficient,yang20233dhumangan}. To manipulate the camera-to-subject distance, they first estimate an initial head model and camera pose and intrinsics that are jointly optimized through the GAN inversion technique \cite{roich2022pivotal} using 
a combination of the photometric-based loss LPIPS \cite{zhang2018perceptual} and the L2 loss between
the landmark detected on the generated and input face to find the camera parameters, face
identity and expressions (\ie the latent face code) that when provided to the generator produce the distorted input image. The generator is then finetuned (overfit) on this selfie input until convergence. The inversion process is time-consuming as the authors of Disco \cite{wang2023disco} report that this operation requires more than 2 minutes on a modern GPU. In addition, their pipeline produces subtle-to-noticeable identity changes, depending on the input, due to the inaccuracies of the encoded face from the inversion of 3D GAN outputs. Moreover, similar to \cite{yin2021learning}, their methods require
cropping the face from the body leading to additional complex post-processing steps to combine the face with the rest of the body
and the background image afterwards.

In \cite{karpikova2024super}, they extend Disco \cite{wang2023disco} by estimating initial camera parameters and the face latent code using pretrained networks to decrease the number of iterations in the inversion process. 

The work of \cite{Chen_2024_CVPR} did not mainly address the task of rectifying perspective distortion, but they were performing it as a subtask of their pipeline which aims to synthesize a selfie photo of the whole body from a group of close-range shots of different body parts.  As the face forms a small portion of the final synthesized image, the quality of the corrected face was not the main target of the method. They follow a flow-based warping method similar to LPUP \cite{yin2021learning} and train the architecture from \cite{wang2021one} on image pairs generated by the 3D GAN network EG3D \cite{chan2022efficient}.  However, the artefacts in the generated training data led to inferior results compared to previous methods.

In \cite{kao2023toward}, they study the problem of 3D face reconstruction under perspective distortion and propose a method that considers a perspective camera model to encounter the problem of distortion instead of the orthogonal model, However, their aim is not to produce the corrected face but rather to reconstruct a consistent 3D mesh under different camera distances for the same face.

The method in \cite{gao2020portrait} is based on neural radiance field representations and adopts a meta-learning approach to generalize to unseen samples at test time. They assume a specific pose about the input (frontal view of the subject). Because their training views are taken from a single camera distance, vanilla NeRF rendering requires inference on the world coordinates outside the training coordinates leading to artefacts when the camera is too far or too close. Furthermore, due to their reliance on volume rendering in NeRF, their method is prohibitively slow during runtime.

\section{Method}

\subsection{Overview}
\noindent Our perspective distortion rectification pipeline consists of the following five main modules as shown in Figure \ref{fig:Arch}: depth estimation, feature extraction, horizontal translation regression, differentiable reprojection and a generation network. Given a selfie or a portrait image and a foreground segmentation mask, which can be acquired by an off-the-shelf human segmentation network \cite{chen2022robust}, we estimate a depth map and extract image features using the depth and feature networks respectively. The predicted depth is used to un-project the 2D image features to the 3D coordinates of the original camera. We virtually manipulate the camera-to-person distance by an offset $t_z$ and increase the focal length proportionally to retain the same face scale as in the input image. Instead of moving the camera straight to the back, we utilize another network to predict a horizontal camera translation $t_x$ from the selfie input. Thus, the camera can move with an angle to the back. This can boost the results of the inpainting module by moving the camera in a direction that decreases the region that needs to be inpainted. The features are then reprojected to the perspective of the new view.  Similarly, the input foreground RGB image and the mask are warped to the new perspective. The reprojected features, RGB image and mask are fed to the generator module, which fills in the holes and missing regions occluded in the original view, refines local artefacts, and outputs a new mask that follows the corrected face. The new mask is used to combine the background (BG) and foreground (FG) images afterwards. Before combining, the BG image is inpainted by another pre-trained network.

\subsection{Depth Prediction}
\noindent We use a residual UNet architecture similar to many recent monocular depth estimation methods \cite{yin2021learning} for facial depth prediction. The encoder and decoder consist of 5 residual blocks for each. The input to the depth network is the segmented foreground human. Hence, the network only estimates the depth over the foreground region.

\subsection{Feature Extraction} 
\noindent Instead of directly reprojecting the image in the RGB space, we extract high-level features from the foreground-segmented input image with another UNet-based architecture composed of 5 mirrored pairs of convolution layers. The feature extractor provides an enriched representation of the underlying RGB image to the generation net. 
It produces a feature map that has the same spatial size as the input but with 64 channels. Thus, every depth pixel is attached to a richer feature vector of length 64.

\subsection{Image Warping}
\noindent Once we have the depth map of the subject, we are able to synthesize a novel view of the subject by manipulating the camera-to-subject distance and the focal length.  We unproject the face to the 3D space of the original camera $C_1$ that has camera intrinsics matrix $K_1$ expressed as follows:
\begin{equation}
 K_1 = \begin{bmatrix}
f_1 & 0 & c_x\\
0 & f_1 & c_y \\
0 & 0 & 1
\end{bmatrix}
\label{eq:camera_matrix}
\end{equation}
where $f_1$ is the focal length of $C_1$ and $c_x$ and $c_y$ is the principal point in x and y directions, respectively. The 3D points are then transformed to the coordinates of a chosen new virtual camera $C_2$, and projected to its perspective using the chosen intrinsics $K_2$ of $C_2$. Given a pixel point $p_1$ in an image captured by $C_1$, The new position $p_2$ of the point $p_1$ in the $C_2$ image plane can be computed as the following:

\begin{equation}
 p_2 = K_2 [R|T] d K_1^{-1} p_1
\end{equation}
where $d$ is the depth of $p_1$, $ \begin{bsmallmatrix} R
 &T \\ 0 & 1 \end{bsmallmatrix} \in {\rm I\!R}^{4\times4}$ is the relative transformation matrix between $C_1$ and $C_2$ composed of the relative rotation matrix $R \in {\rm I\!R}^{3\times3}$ and translation vector $T \in {\rm I\!R}^{3}$. The transformation to and from homogeneous coordinates has been omitted for brevity. 
Because (2) is differentiable, it can be directly plugged into the network training.% But, now we need to render the new view image (\ie warping the image). %This equation is composed of three steps. First, we unproject the 32d image points to the 3D coordinates of the first camera by computing dk-1 p1, Second transform the point cloud from the first camera to the second camera using the relative transformation matrix p3d2 = [R]p3d1, and last project the 3d point cloud into the perspective of the camera matrix using  and remove the scale by dividing by the z (depth) value  

Rendering the actual image (\ie warping the image) is called the rasterization process. To enable E2E training of our model, we leverage a differentiable renderer module of the PyTorch3D framework \cite{ravi2020pytorch3d}.

\begin{figure*}
\centering

\includegraphics[width=1.0\textwidth]{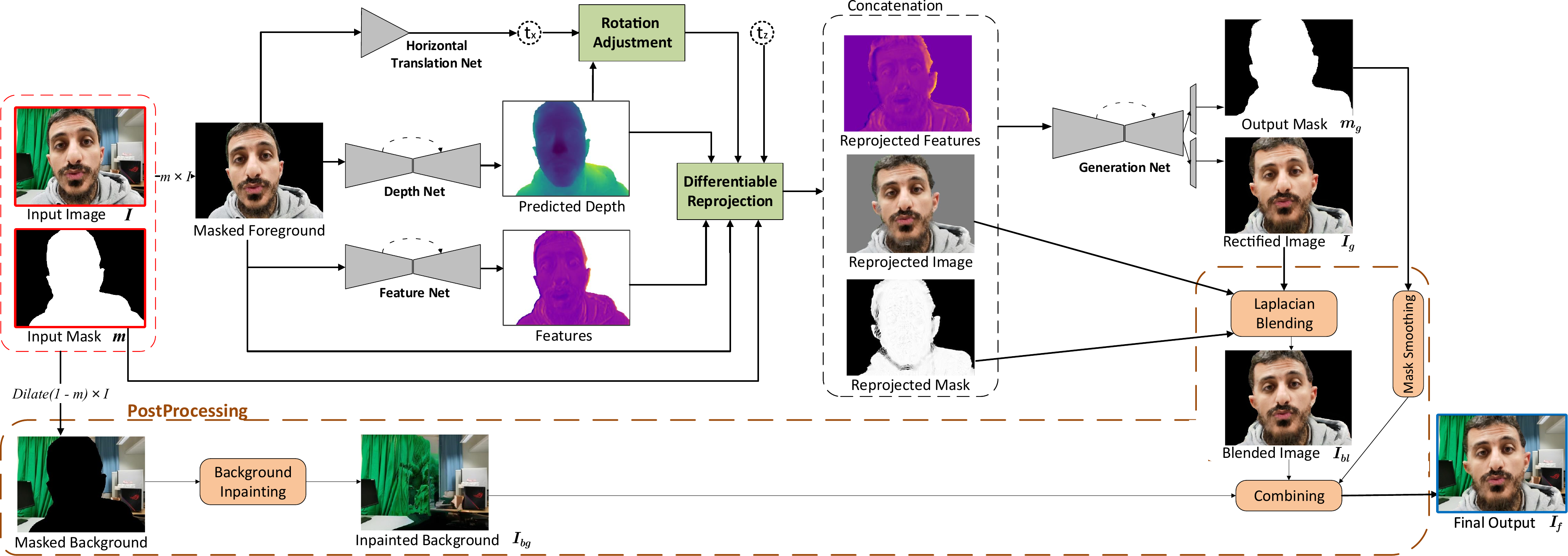}

\caption{Architecture of our end-to-end  perspective rectification pipeline. The postprocessing steps are enclosed by an orange dashed line. The number of channels of the feature maps is 64 but only the first feature map is visualized here for illustration.  }

\label{fig:Arch}
\end{figure*}

\subsection{Camera Horizontal Translation}
\label{sec:trans}
\begin{figure}[h]

 \includegraphics[width=1.0\linewidth]{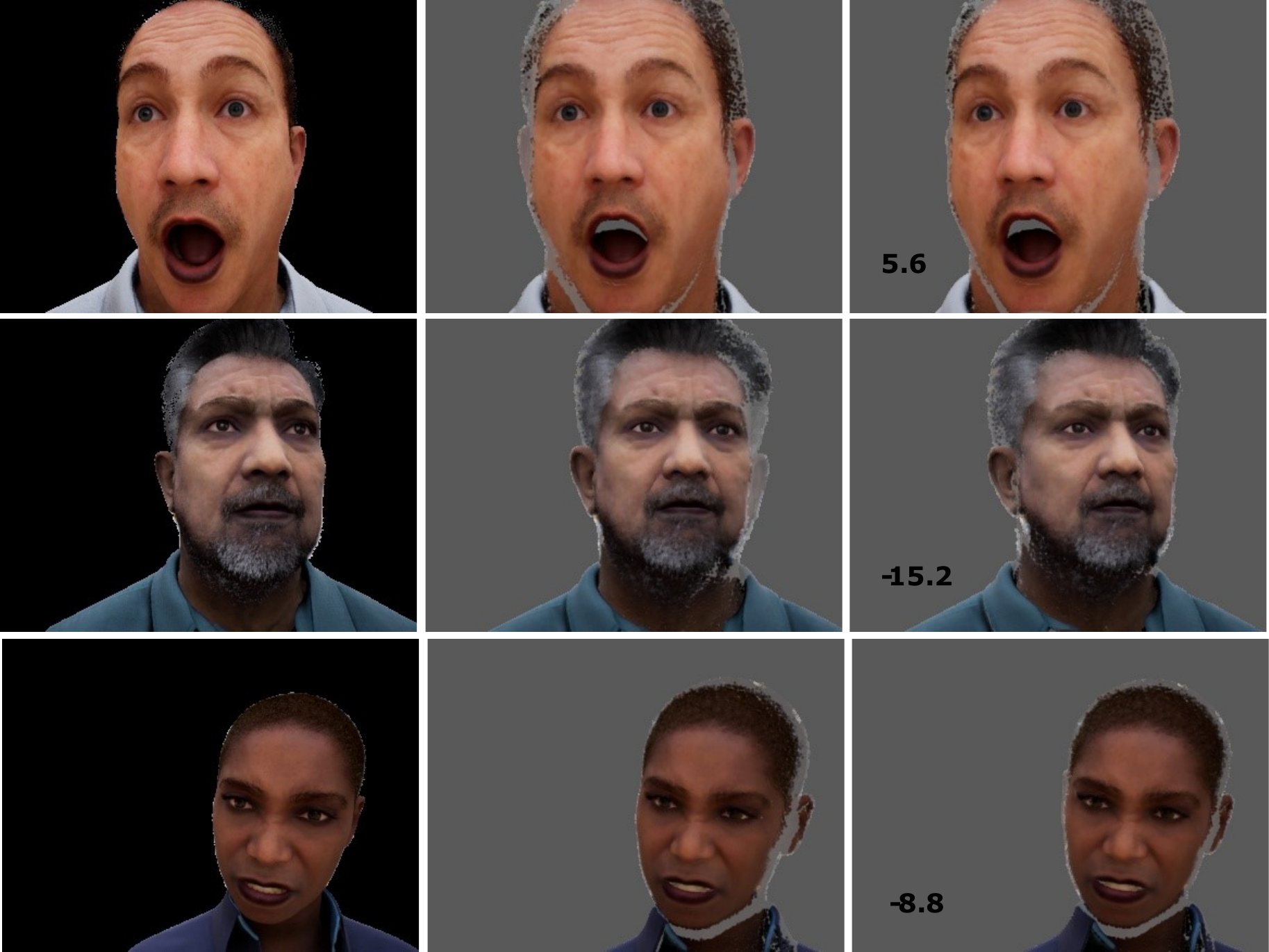}
\begin{tabularx}
{1\linewidth} { 
   >{\centering\arraybackslash}X 
   >{\centering\arraybackslash}X 
   >{\centering\arraybackslash}X}
 Selfie input & $t_x = 0 $ & $t_x \neq 0 $  \\

\end{tabularx}

\caption{Illustration of the difference between moving the camera with horizontal translation $t_x\neq0$ and without $t_x=0$ in terms of the missing region. The warped input in both cases is overlaid on the GT target distortion-free image. The translation value is indicated over each image in cm. The negative value means a translation in the direction of the left-hand side. In each example, we show a camera horizontal translation that decreases the missing face region.}

\label{fig:horiz_missing}
\end{figure}

\noindent The question that arises then is how does the camera move to the back; does it move straight or with an angle? In Figure \ref{fig:horiz_missing},  we show an example where the camera moves back straight versus the case where the camera moves with an angle. The warped image is laid over the GT target image to show the missing area after warping the input image. It is shown that moving the camera to the back along the optical axis can have more missing details to be predicted than the movement with some angle. To improve the quality of the synthesized images, we introduce an auxiliary module that takes the distorted image as input and predicts the horizontal camera translation that if applied would decrease the area that needs to be hallucinated by the inpainting module.
Now the camera has a translation along the Z-axis $t_z$ (\ie along the optical axis) and a horizontal translation along the X-axis $t_x$ relative to the original position. Hence, the translation vector of the camera becomes $T = [t_x, 0, t_z]^T$ relative to the original camera $C_1$. When the camera moves only backwards, $t_x$ equals zero. Introducing a horizontal translation would shift the face position in the warped image. So, we need to compute a rotation angle $\theta$ around the Y-axis of the camera to retain the warped face in the same position as the input image (or the position of the face if the camera moved back along the optical axis, \ie $t_x=0$). Let's assume that the camera that moves back along the optical axis is $C_2$ and the camera that moves to the back with horizontal translation $t_x \neq 0$ and rotation $\theta$ is $C'_2$. 
The relative transformation matrix from $C_2$ to $C'_2$  is :
\begin{equation}
    M = \begin{bmatrix} R' & T'\\ 0 & 1 \end{bmatrix}   = 
    \begin{bmatrix} 
    cos{\theta} & 0 & sin{\theta} & t_{x} \\ 
    0 & 1 & 0 & 0  \\ 
    -sin{\theta} & 0 &cos{\theta} &0  \\  
    0 & 0 & 0 & 1 
    \end{bmatrix} 
\end{equation}

To compute the rotation $\theta$, we choose an arbitrary point in the face where this point after translating and rotating the camera by $t_x$  and $\theta$ respectively, should have the same position as if we moved the camera back with $t_x$ and $\theta$ are zeros.  We choose this pixel point on the left face boundary if the subject is closer to the left image side and vice versa. Choosing the point on face boundaries guarantees that this point will not go outside the image boundary after the reprojection. This is especially important when the face is close to the image boundaries; otherwise, any point is valid. Let's suppose the chosen pixel point $p$ in the input image is at position $(u,v)$ expressed in 2D image coordinates. The 3D coordinates of the point $p$ in camera $C_2$ is $P_{2}$ = $[X_2,Y_2,Z_2]^T$. The transformation of $P_{2}$ to the coordinates of camera $C'_2$ is:
\begin{equation}
P'_{2} = \begin{bmatrix}
    X'_2 \\  Y'_2  \\ Z'_2
\end{bmatrix}=M^{-1} P_{2}=
    \begin{bmatrix} X_2 cos{\theta} - Z_2 sin{\theta} - t_X cos{\theta} \\ Y_2 \\ 
    X_2 sin{\theta} + Z_2 cos{\theta} - t_X sin{\theta}  
    \end{bmatrix}
\label{eq:mult}
\end{equation}
and we have $u = f \frac{X_2}{Z_2}  + c_x$, and we want the pixel position $u$ in the image formed by $C_2$ and $u'$ in the image formed by $C'_2$ to be equal, Thus: 
\begin{equation}
   \frac{X_2}{Z_2} = \frac{X'_2}{Z'_2}
   \label{eq:ratio}
\end{equation}

By substituting (\ref{eq:mult}) into (\ref{eq:ratio}) , we get: 

\begin{equation}
   \theta  = arctan(\frac{-t_x Z_2}{Z_2^2 + X_2^2  - t_x X_2})
   \label{eq:theta}
\end{equation}
We use a ResNet18 \cite{he2016deep} architecture for our translation module. It predicts a vector $V$ of length $l= 50$ which corresponds to the equal-spacing discretization of the translation range $[-20,20]$ cm.
% equal-width discretization, uniform 
During inference, we find the translation  by computing the Softmax of the output vector as the following:
\begin{equation}
    \hat{t}_x = \sum_{j} t_x^j \cdot \text{softmax}(\hat{V}(j)) 
    \label{eq:trans}
\end{equation} 
where $t_x^j$ is the horizontal translation at bin $j$.

\subsection{Image Refinement}
\label{sec:img_refin}
\noindent The warped image contains holes and missing pixels which were occluded in the original view.  Unlike regular inpainting tasks where the missing regions are given as input, the missing pixels on the boundary of the face are not known to the network in advance. Thus, the network should learn to predict where to inpaint along with the hallucination of the content. In other words, it needs to learn which pixels are actually missing and belong to the foreground (\ie the face) and which belong to the background.  
Moreover, the network has to fill in incomplete pixels (\ie pixels that exist partially) and fix local errors resulting from the warping process.

The generator network has a similar architecture as the feature extractor except in the input and output layers. The input is the concatenation of warped features from the extractor, warped foreground RGB image and the warped input mask. The warped mask ranges from 0 to 1 indicating whether the pixel is missing, incomplete or exists. The generator outputs the completed face and a foreground mask that follows the foreground output image. This mask is used to combine the foreground and background afterwards.

\subsection{Post Processing and Image Composition}
\noindent Overall, our pipeline contains three main postprocessing operations as illustrated in Figure \ref{fig:Arch}: Laplacian blending, background inpainting, and image combining. First, the generator output is blended with the warped RGB image through Laplacian pyramids \cite{adelson1984pyramid} to leverage the high-frequency details of the warped image and produce more plausible and cleaner results. Separately, we use the off-the-shelf SOTA inpainting network MAT \cite{li2022mat} to fill in the missing parts of the background image. Before BG inpainting, the input mask is morphologically dilated to remove any remains of the FG from the BG. Finally, for the blended FG and inpainted BG composition, we smooth the sharp boundaries of the generator output mask with a Gaussian filter.  We then combine the two images as the following: $I_f =  (1-m_s) \cdot I_{bg} + m_s \cdot I_{bl}$ where $m_s$ is the smoothed mask and $I_{bg}$, $I_{bl}$ and $I_f$ are the inpainted BG, the blended FG and the final composed image, respectively.

\subsection{Training and Loss Functions}
\label{subsec:Training_and_loss}

\noindent \textbf{Depth Loss.} To train the depth network, we use a loss function that is typically used in training monocular depth estimation networks \cite{yin2021learning}. Specifically, we use L1 loss (\ie the mean absolute error) $\fl_{L1}$ between the predicted depth and ground truth depth. We also compute $\fl_{tanh}$ which is the L1 loss between the hyperbolic tangent (\ie $tanh$ function) of each of them, and $\fl_{grad}$ which is the multi-scale gradients loss \cite{yin2021learning}. The overall loss for depth training is then defined as the following:

\begin{equation}
    \fl_{depth} = \alpha_1 \fl_{L1} + \alpha_2 \fl_{tanh} + \alpha_3\fl_{grad}
    \label{eq:dep_loss}
\end{equation}
where the alphas represent the weight of the individual losses.

\textbf{End to End Loss.} 
For E2E training, we use photometric losses and GAN-based losses. For the photometric loss $\fl_{photo}$, L1 loss and perceptual loss $\fl_{pcp}$ \cite{johnson2016perceptual} are computed between the generated image $I_g$ and the ground truth target image $I_t$ over three regions: the whole image, face region, and missing pixels.  $\fl_{photo}$ is defined as the following: 
\begin{equation*}
   \fl_{photo} = \sum_{i}\omega_i \fl_{photo}^{i} \text{ \,}\quad i  \in \{all, face, missing\}, 
\end{equation*}
\begin{equation}
     \text{where} \quad \fl^i_{photo} =  \mu \fl^i_{L1} +\gamma \fl^i_{pcp}
    \label{eq:photometric_loss_overall}
\end{equation}

The GAN-based training loss \cite{wang2018high} for the generator is defined as follows:
\begin{equation}
    \fl_g = \emean[D_f(I_g)-D_f(I_t)] - \emean[D(I_g)]
\end{equation}
where $D(.)$ is the discriminator network output where values $\leq -1$ represent fake images and  values $\geq 1$ mean real ones. $D_f(.)$ is the feature map from the discriminator and $\emean(.)$ computes the mean over the elements.

The overall loss is as follows: 
\begin{equation}
    \fl_{overall} =  \fl_{depth} + \fl_{photo} + \beta \fl_g + \rho \fl_{mask}
    \label{eq:overall}
\end{equation}
where $\fl_{mask}$ is the binary cross entropy loss between the predicted mask from the generator $m_g$ and the GT mask $m_t$. $\beta$ and $\rho$ are the loss weights.

For the discriminator training, we use the hinge loss \cite{wang2018high} which is defined as the following:
\begin{equation}
    \fl_D = - \emean[min(0,-1-D(I_g))] - \emean[min(0,1-D(I_t))] 
    \label{eq:disc_loss}
\end{equation}

\textbf{Horizontal Translation Module Training and Loss.} 
During our experiments, we found it hard to train the translation network (HzT net) in an unsupervised way by incorporating it directly into the pipeline. Thus, we generated the training GT targets to train the network in a supervised way. To compute the GT translations, we discretize the possible range of camera horizontal translation, as mentioned in Section \ref{sec:trans},  to 50 translation steps or bins. For each step,  we compute the corresponding rotation according to (\ref{eq:theta}), warp the image to the new pose and provide it to our pretrained generator module which predicts the rectified image and mask as explained in Section \ref{sec:img_refin}. We compare the predicted output mask $m_g$ with the GT target mask $m_t$. The details of synthesizing $m_t$ are explained in Section  \ref{sec:Data_Generation}. We use the intersection over union (IoU) to compute the difference between $m_g$ and $m_t$ as the following:

\begin{equation}
    IoU = \frac{\sum_{i=1}^{N} {m_g(i) \cdot m_{t}(i)}} {\sum_{i=1}^{N} {m_g(i) + m_{t}(i) - m_g(i) \cdot m_{t}(i)}}
    \label{eq:IoU_loss}
\end{equation}
where $N$ is the number of pixels in the mask. We then assign a value $1$ to the bin with the maximum $IoU$  value. The remaining bins take values according to the following rule:

\begin{equation}
V(j) = 
\begin{cases}
    1 & \quad \text{if \,}   \delta_j <= 0.01 \\
    0.9 & \quad \text{if \,} 0.01 < \delta_j <= 0.02\\
    0 & \quad \text{if \,} \delta_j > 0.02

\end{cases}
\end{equation}
where $\delta_j$ is the relative difference between the $IoU$ value for a bin $j$, and the maximum $IoU$, as follows: 
${\delta_j = (IoU_{max} - {IoU}_{j})} {/ (1-{IoU}_{max}})$.

By following these steps, we have a target vector $V$ that will be used for training our translation module.

We use the binary cross-entropy loss between the predicted vector $\hat{V}$ from the HzT net and the target vector $V$. The predicted translation is estimated according to (\ref{eq:trans}) and used to warp the input foreground mask $m$. We then compute the IoU loss between the warped mask $m_w$ and $m_t$. The total loss for training the HzT module becomes as the following: 
\begin{equation}
    \fl_H = \fl_{BCE}(\hat{V},V) + \lambda \fl_{IoU}(m_w,m_t)
    \label{eq:trans_loss}
\end{equation}
where $\fl_{IoU}(m_w, m_t) = 1- IoU(m_w, m_t) $, and $\lambda$ is a hyperparameter for weighing  $\fl_{IoU}$. The first term encourages the network to focus on the pitfalls of the inpainting modules since the target vector $V$ is based on comparing the mask predicted from the generator $m_g$ with the GT mask $m_t$, while the second term encourages the network to find the translation that minimizes the missing region in general apart from its difficulty to the inpainting network.

\section{Dataset Generation}
\label{sec:Data_Generation}
%synthetic data generation 
\noindent To train our model, we need pairs of perspective-distorted images and their distortion-free targets. In \cite{zhao2019learning}, they used the BU-4DFE facial dataset \cite{zhang2013high} for generating training pairs. This dataset has static illumination and expressions for each of the subjects. The 3D head is reconstructed from the synchronized multiple cameras. The reconstruction does not include the subject's body and usually has limited variations of head poses.  The distorted version of the faces are rendered simply using perspective projection and rasterization, without changing illumination or considering complex reflectance properties. Unlike \cite{zhao2019learning}, we utilize Unreal Engine (UE) for rendering a realistic synthetic facial dataset.  UE offers realistic human characters called Metahuman (MH). Those MH are captured in different locations of a simulated city project in UE  called the City Sample. During capturing, we also vary the camera-to-subject distance, the camera orientations, the head pitch, yaw, roll and facial expressions. 

\begin{figure}

\includegraphics[width=1.0\linewidth]{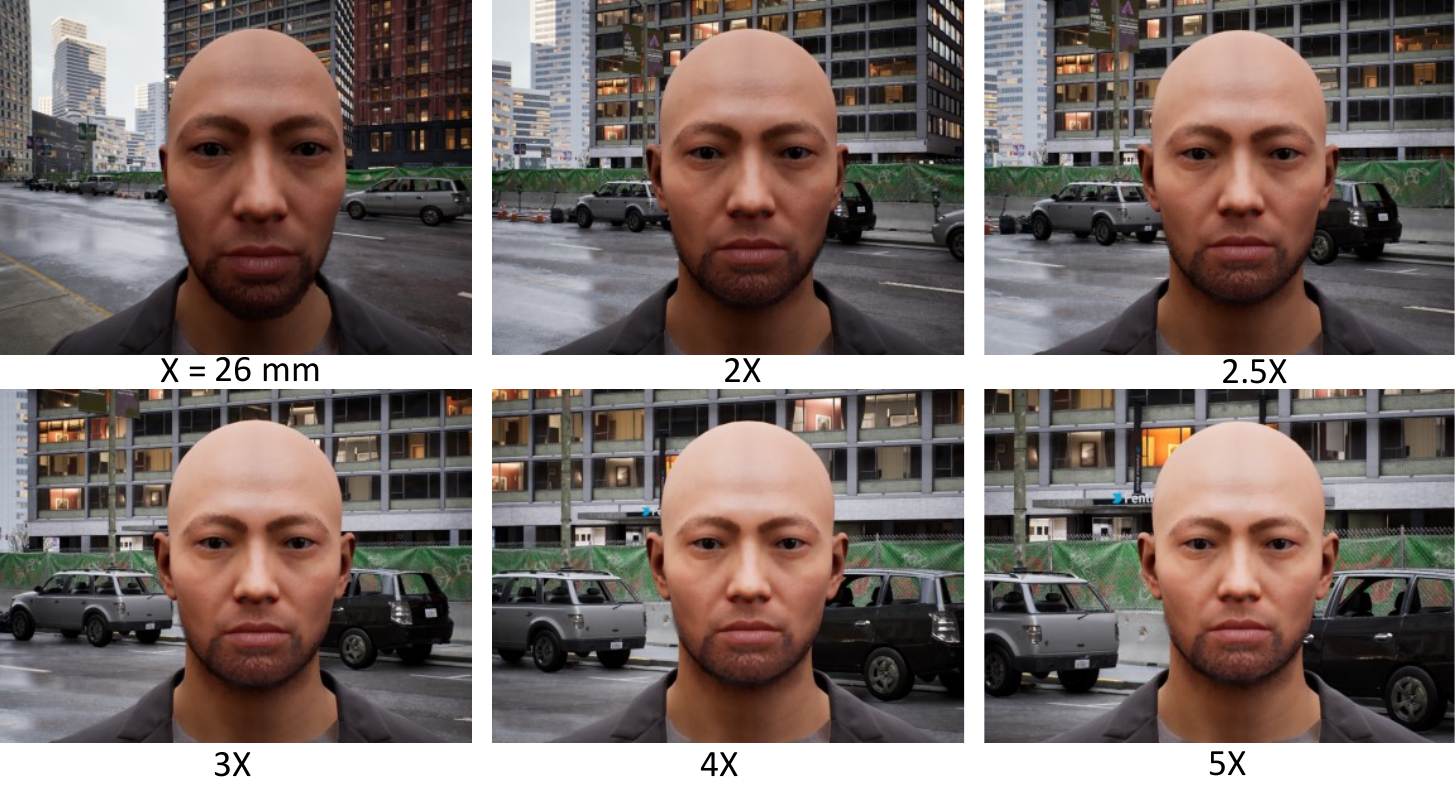}

\caption{An example of a synthetic human face captured in UE by a camera with a focal length of 26mm which has perspective distortion. The focal length is varied while increasing the subject-to-camera distance proportionally. The 3X focal length is reasonable to represent the distortion-free target image of the input distorted image and is similar to the larger focal lengths.}

\label{fig:X_to_5X}
\end{figure}

The close-range camera in our simulation has a 35mm-equivalent focal length of 26mm which is similar to the typical focal length of selfie or wide-angle cameras in nowadays' smartphones. The far-range camera has 3x optical zoom relative to the selfie camera. We found that 3x optical zoom shows an acceptable and plausible perspective projection of the face and has minimal differences over the focal lengths beyond 3x as shown in Figure \ref{fig:X_to_5X}. In addition to the generated images from both cameras, the human subject depth map and mask are generated as well.

We used two types of MH characters: the first is a predefined set of characters provided by UE, and the second is a subclass instance of the MH character called `CitysSampleCrowd' character. The latter is a pool of different faces, outfits, skin textures, and grooms (\ie hair, beard, moustache, etc). The character is then randomly generated from those various options.  %offering versatile identities and outfits.
In the first type, we used 66 different characters, from which 10 characters are dedicated for testing. To simulate individuals wearing eyeglasses, we utilize a UE asset package containing 18 different types of sunglasses and reading glasses for both men and women.

\begin{figure}[h]

 \includegraphics[width=1.0\linewidth]{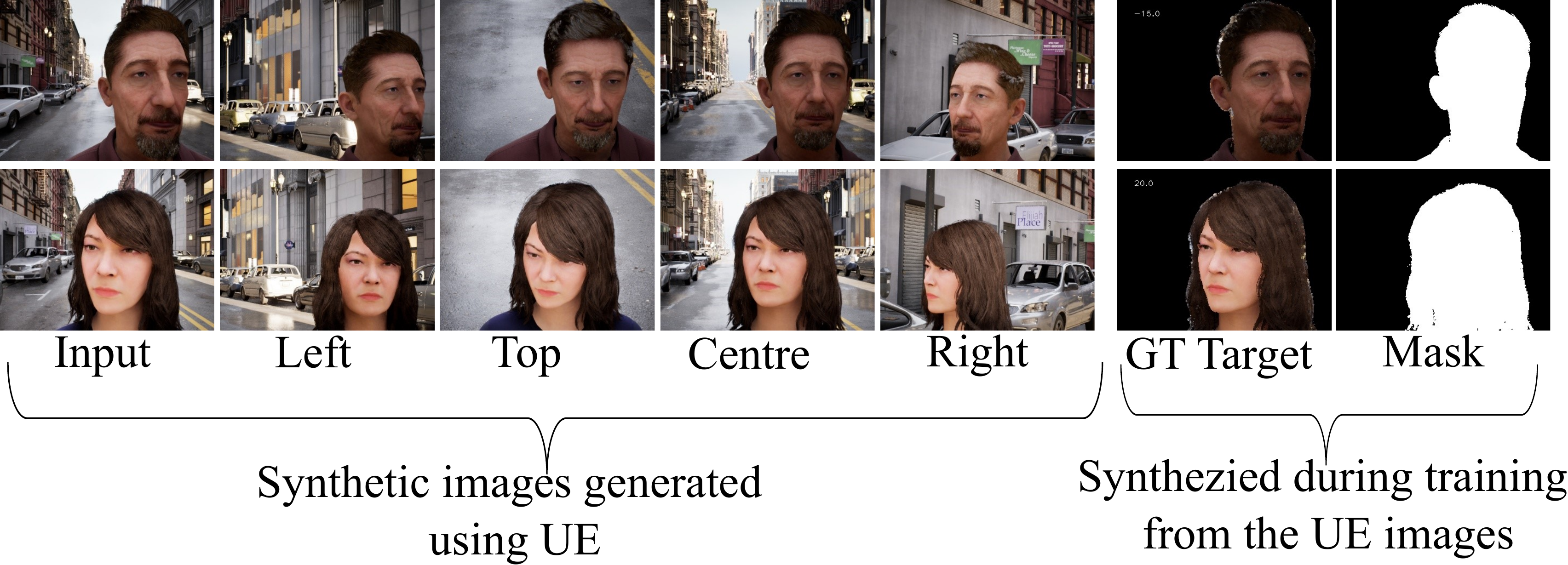}

\caption{Two examples showing two persons from the multiview dataset. The input represents the close-range image while the left, top, centre and right view images represent the far-range image (3X focal length camera) from different angles relative to the front wide-angle camera. During the Horizontal translation module training, we synthesize the GT target image and mask from the multi-view far-range images with the predicted translation $t_x$. The $t_x$ value used in those examples is indicated over each GT target. }
\label{fig:Synthesizing_multiview_example}
\end{figure}

 We generated two sets of synthetic data: a single-view and a multi-view dataset. In the former, the close-range and its corresponding far-range images are generated. In the multi-view case, we capture multiple far-range views. Specifically, we generate 4 rear views: left, right, centre, and top side of the front camera, as illustrated in Figure \ref{fig:Synthesizing_multiview_example}. The poses of those cameras are fixed relative to the front camera. This dataset is used for the training and evaluation of the horizontal translation module as explained in Section \ref{subsec:Training_and_loss}. 
 
 We generated around 42K single-view samples with resolution $800\times600$ (around 9K samples were originally generated with resolution $1600\times1200$). The synthetic dataset contains around 4.2K image pairs with subjects wearing various eyeglasses which represent $\sim 10\%$ of the dataset. It took around 10 days to generate those samples on a machine equipped with a 12-core AMD Ryzen5900X CPU processor and a NVIDIA GeForce RTX 3090 GPU. The multi-view training and testing dataset has around $4\times2000$ and $4\times1000$ images, respectively, with original resolution $800\times600$. 

 To train the Horizontal translation module described in Section \ref{sec:trans}, we need to synthesize the ground-truth target image on the fly during training since we have only the view that corresponds to $t_x=0$ (\ie the camera moves straight to the back). To synthesize the target image with $t_x\neq0$, we reproject the view of the rear centre camera to the target pose. Similarly, we warp the other rear views to the same target pose. The missing pixels from the warped centre view are filled from the other reprojected rear views. The synthesized target mask is used during training, while the target colour photo is only used to report the photometric errors.

\section{Experiments and Results}
 
\subsection{Training details} 
\noindent We first finetune a pretrained monocular depth estimation model from \cite{yin2021learning} till convergence for 50 epochs using our synthetic facial dataset. Inferring the depth of a facial image directly by a pretrained depth regressor without finetuning on our dataset produces an almost flat depth map for the subject. The loss function is computed according to (\ref{eq:dep_loss}), where all weights are set to $1.0$ except $\alpha_2$ is set to $0.5$. The loss is only computed within the foreground region using the mask.
The learning rate is set to $10^{-4}$ and decreased by 0.5 after 20, 30 and 40 epochs. The stochastic gradient descent is used as an optimizer following \cite{yin2021learning}.

Next, we pretrain the feature extractor ($E$) and generator ($G$) nets shown in Figure \ref{fig:Arch} where the ground truth depth is used in projection instead of the predicted one. The loss function is computed according to (\ref{eq:overall}) where $\fl_{depth}$ is disabled, and  $\omega_{all}$, $\omega_{face}$, $\omega_{missing}$, $\beta$ and $\rho$ are set to $0.2$, $1$, $5$, $1$ and $20$, respectively. Adam optimizer \cite{kingma2014adam} is used and the learning rate is $5 \times 10^{-4}$ and halved every 50 epochs with a total of 200 epochs.

After that, we train the depth network in E2E fashion while freezing the $E$ and $G$ nets. The loss function at this stage is computed according to (\ref{eq:overall}) while setting $\omega_{missing}$ and $\beta$  to zero, and $\rho$ is set to $10$. The depth loss weights are set as in the first stage. This kind of scheme offers stability in the training process. 
The horizontal translation module is trained separately using the synthetic multiview dataset for 50 epochs with a learning rate of $10^{-4}$. The $\lambda$ in (\ref{eq:trans_loss}) is set to $0.5$.

Finally, we finetune the $E$ and $G$ nets while the depth net is frozen and its output is used for reprojection. This enables the $G$ net to adapt to the nature of the holes produced by warping the image features using the estimated depth and refine its errors.  At this stage, the loss function is computed according to (\ref{eq:overall}) where $\fl_{depth}$ is disabled.

%evaluation metrics
\subsection{Evaluation metrics}
\noindent We utilize the following metrics to evaluate the performance of our perspective rectification solution: 

\begin{itemize}
    \item  Photometric errors between the final rectified output and the GT distortion-free image such as peak signal-to-noise ratio (PSNR) and structural similarity (SSIM) \cite{wang2004image} and the learnt perceptual similarity metric LPIPS \cite{zhang2018perceptual}.
    \item  Euclidean distance between the estimated facial landmarks of the predicted output and the reference image. We use Mediapipe \cite{lugaresi2019mediapipe} to estimate dense facial landmarks and use them to align both images by a similarity transform. We then compute the Euclidean distance between the corresponding landmarks.  
\end{itemize}

\begin{table}[h]
    \centering
\caption{The results of our pipeline versus the LPUP approach. The "Before" case reports the similarity measure between the distorted input and the ground truth distortion-free image. The arrows indicate whether higher `$\uparrow$'  or lower `$\downarrow$' is better for a metric.}

\label{tab:quantitative_results_e2e}
\begin{tabularx}{0.5\textwidth} { 
  l
  | >{\centering\arraybackslash}X 
  | >{\centering\arraybackslash}X   }
         \hline 
         Method&  PSNR $\uparrow$  &SSIM $\uparrow$ \\ \hline
         Before   & 14.05 & 0.4990 \\
         LPUP \cite{yin2021learning} & 17.80   &0.6051  \\
         Ours & \textbf{19.89}  &  \textbf{0.8332}\\ 
 \hline
\end{tabularx}
\end{table}

\begin{figure}[h]
\centering

 \includegraphics[width=1.0\linewidth]{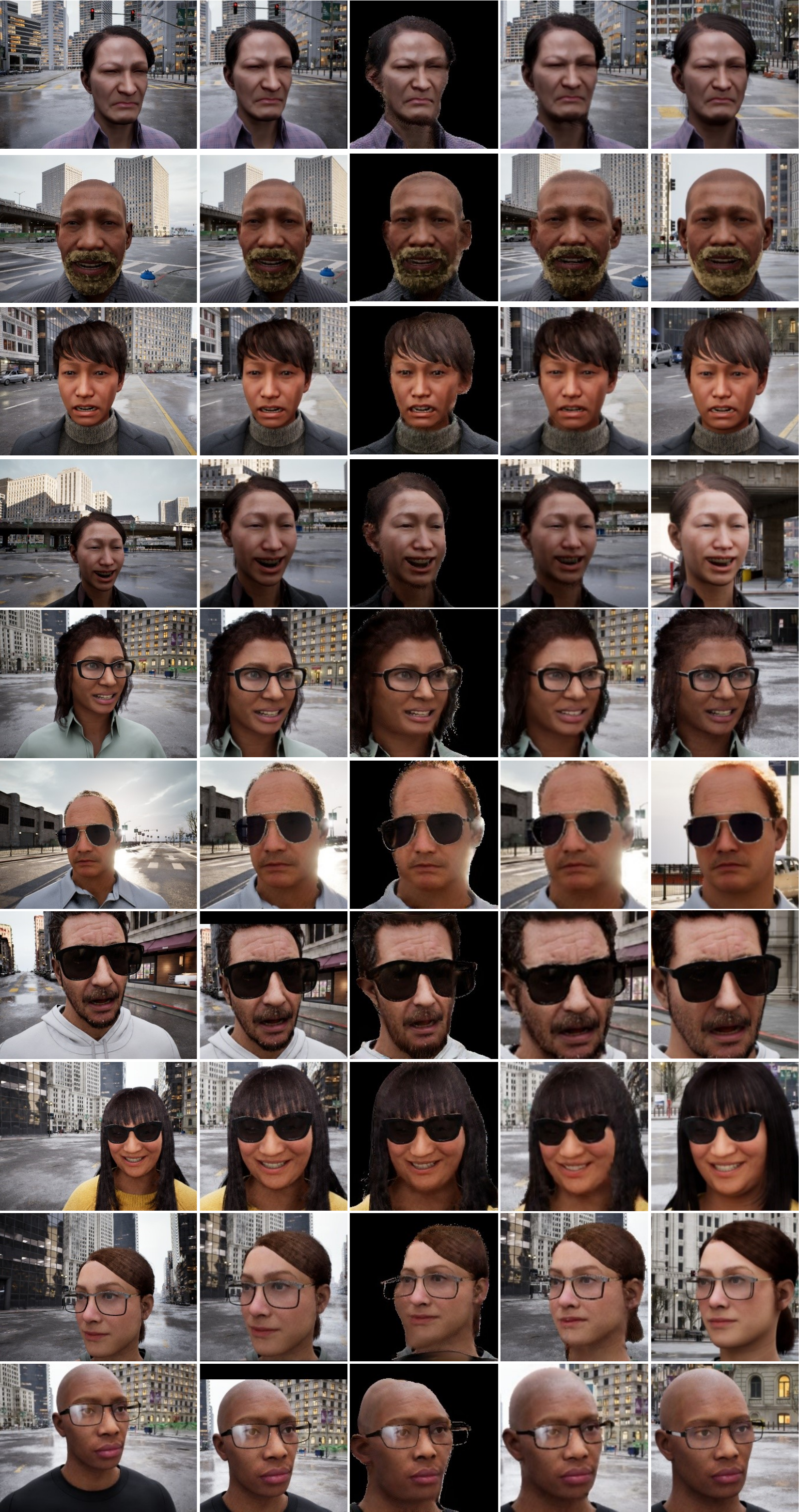}

\begin{tabular}{m{1.8cm}  m{1.2cm}  m{1.5cm} m{1.3cm} m{1cm} } 
\centering
 Input & rescaled & LPUP \cite{zhao2019learning} & Ours &  GT \\ 
\end{tabular}

\caption{Qualitative comparison between our method and LPUP \cite{zhao2019learning} on the synthetic dataset. Our method takes the full frame input as input while LPUP \cite{zhao2019learning} takes a cropped and rescaled image as input. For easier comparison, we rescaled our results to match the size of the LPUP input. The background of the input is not warped, therefore it does not correspond to the background of the ground truth image.}

\label{fig:Ours_vs_LPUP_synthetic}
\end{figure}

\subsection{Results} 
\noindent We compare our pipeline to the following previous portrait correction solutions: Fried et al. \cite{fried2016perspective}, LPUP \cite{zhao2019learning}, DFN \cite{nagano2019deep}, and Disco \cite{wang2023disco}. Additionally, we compare our method to Shih's method \cite{shih2019distortion} which is a solution for correcting a different type of face distortion that occurs in images captured with wide-angle cameras when the subject is close to the edges of the image, which is solved by stereographic projection. DualPriors (DP) \cite{yao2024combining} is a recent method that addresses the same problem as Shih \cite{shih2019distortion}. 

Since LPUP's authors \cite{zhao2019learning} do not release their code or datasets, we reimplemented the first part of the LPUP pipeline so that we can compare their proposed pipeline with ours. Specifically, we replicate the training of the Flownet part of the pipeline. The input to the network consists of an RGB image and a depth label. The depth label is repeated to have the same spatial size as the input image and concatenated with the image channels. The network output is a two-channel deformation map used to warp the input image to correct the perspective distortion. Second, we add a differentiable bilinear sampling module to make a backward warping of the GT distortion-free image using the predicted and ground-truth flow maps to the perspective of the input selfie image. The warped images are labelled as fake and real, respectively, and used to train a discriminator network. Thus, the loss function for training the flow map network becomes similar to LPUP \cite{zhao2019learning}. For a fair comparison, we use our synthetically generated dataset for the training. The dataset is preprocessed in the same way as in LPUP. Particularly, we scale and crop the image such that all faces in the images have approximately the same scale, and match the inner corner of each subject’s right eye to a fixed position through all images. 

In the qualitative and quantitative comparison with Fried \cite{fried2016perspective}
and Disco \cite{wang2023disco}, we are constrained by the results reported in their paper or project websites, as no codes have been released by either method at the time of paper submission.

Table \ref{tab:quantitative_results_e2e} shows a quantitative comparison between our pipeline and LPUP solution on the synthetic dataset. Our method significantly outperforms LPUP  as indicated by the PSNR and SSIM metrics. 
Although the ground truth depth label is used for the evaluation of LPUP giving it a little bit of an advantage, our method still performs better. In addition to the quantitative results, Figure \ref{fig:Ours_vs_LPUP_synthetic} shows a qualitative comparison on the synthetic dataset.

\begin{table}
    \centering

\caption{Quantitative comparison on the CMDP dataset \cite{burgos2014distance}. LMKE denotes the landmark error, normalized by the image size ($512 \times 512$ pixels). Methods marked with (*) are evaluated on a smaller subset of the CMDP dataset used in \cite{nagano2019deep}.}

\label{tab:cmdp}
\begin{tabularx}{0.5\textwidth} { 
  l
  | >{\centering\arraybackslash}X 
  | >{\centering\arraybackslash}X   }
         \hline 
         Method&  LMKE $\downarrow$  & LPIPS $\downarrow$ \\ \hline
         Input   & 0.00809 & 0.2489 \\
         Fried \cite{fried2016perspective} & 0.00577   & 0.2128   \\
         Disco \cite{wang2023disco} & \textbf{0.00448} & \underline{0.1934 } \\
         Shih \cite{shih2019distortion}&  0.00800 &  0.2665 \\
         DP \cite{yao2024combining}& 0.00838 &  0.4094 \\
         Ours & \underline{0.00453}  &  \textbf{0.1919}\\ 
 \hline

         DFN* \cite{nagano2019deep} & {0.00613}  &  {0.2628}\\ 
         Ours* & { 0.00499}  &  {0.2205}\\ 
         
 \hline
\end{tabularx}
\end{table}

\begin{figure*}[ht]
\centering

 \includegraphics[width=1.0\textwidth]{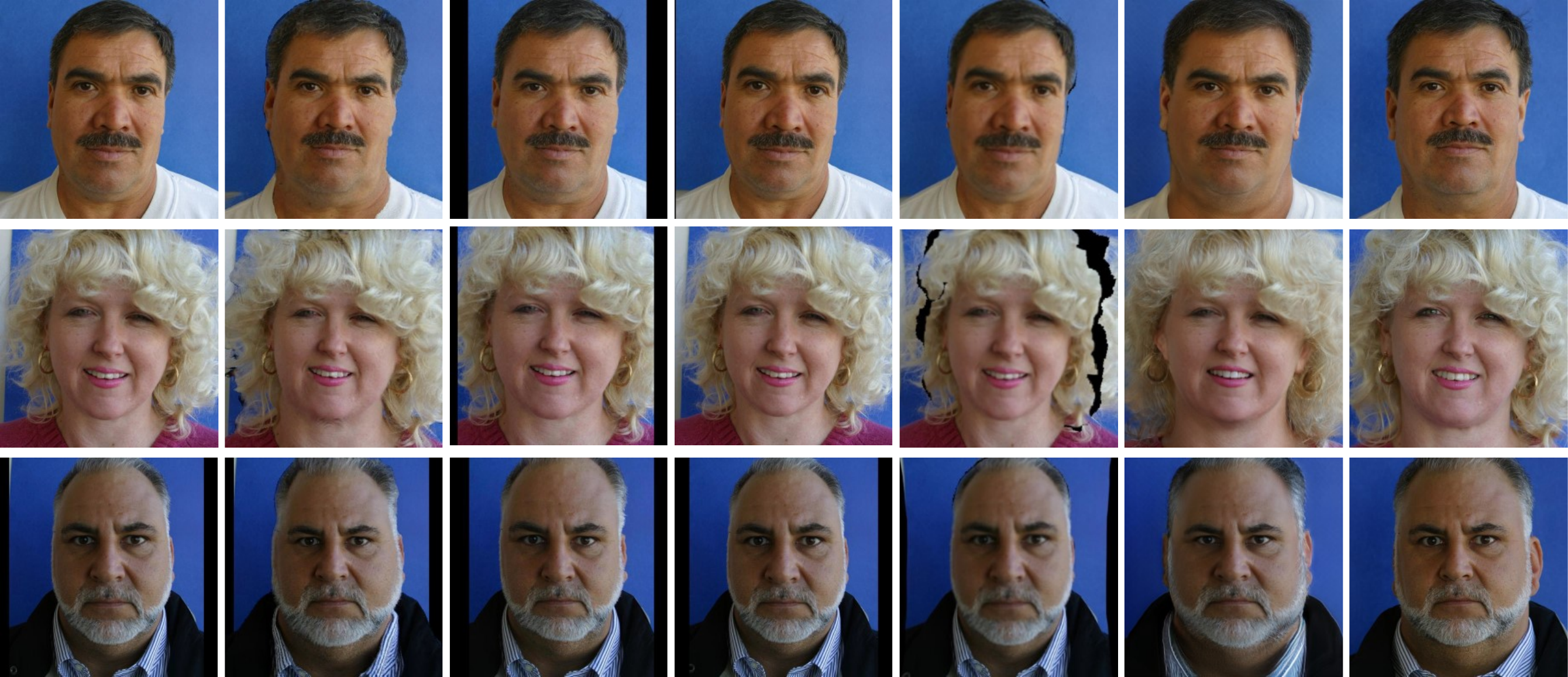}

\begin{tabularx}
{1\textwidth} { 
   >{\centering\arraybackslash}X 
   >{\centering\arraybackslash}X 
   >{\centering\arraybackslash}X 
   >{\centering\arraybackslash}X  
   >{\centering\arraybackslash}X 
   >{\centering\arraybackslash}X
   >{\centering\arraybackslash}X}
 Input & Ours & Fried \cite{fried2016perspective}  & Shih \cite{shih2019distortion} & DP \cite{yao2024combining} & Disco \cite{wang2023disco} &  Reference \\

\end{tabularx}

\caption{Qualitative comparison on the CMDP dataset \cite{burgos2014distance}. The DP method \cite{yao2024combining} produces an image and a mask. The mask can be used to fill the background holes with an external inpainting method. Here, we show the direct output obtained by running their code.}

\label{fig:Comparison_CMDP}
\end{figure*}
We also conduct a comparison on the Caltech Multi-Distance Portraits (CMDP) dataset \cite{burgos2014distance}. This dataset contains portrait images of several subjects captured from various distances. For each identity, the camera-to-subject distance is varied as well. Since the ground truth relative poses between the cameras are not known and the images for the same persons were not synchronized, it is hard to use direct intensity-based metrics such as PSNR or SSIM for measuring the quality. Instead, we estimate the Euclidean distance errors between the aligned facial landmarks of the output image and the pseudo-target image. We use the images captured at distances of 60cm and 480cm as the input and reference, respectively. Using the facial landmarks, we estimate a similarity transform to align the predicted and reference images and compute the perceptual similarity LPIPS on the masked foreground. Table \ref{tab:cmdp}, shows 
 a quantitative comparison on the CMDP dataset. Additionally, Figure \ref{fig:Comparison_CMDP} shows a qualitative comparison on the same dataset. Our visual results look more consistent with the reference image than the correction of Fried \cite{fried2016perspective}, Shih \cite{shih2019distortion} and DP \cite{yao2024combining}, while being comparable with the results of the time-consuming Disco method \cite{wang2023disco}.  Note that Shih \cite{shih2019distortion} and DP \cite{yao2024combining} are not able to rectify perspective distortion in close-range portraits.
 
 The running time of our method on an HD image (1280 $\times$ 720) is 0.4 plus 0.1 seconds for face rectification and background filling, respectively on a machine equipped with an A100 GPU. On the other hand, the running time of the Disco method \cite{wang2023disco} for only the inversion process is approximately 130 seconds on a crop of size $256 \times 256$ as reported by the authors. The time for other postprocessing steps to combine the crop back with the rest of the body is not reported.

\begin{figure*}[t]
\centering
 \includegraphics[width=1.0\textwidth]{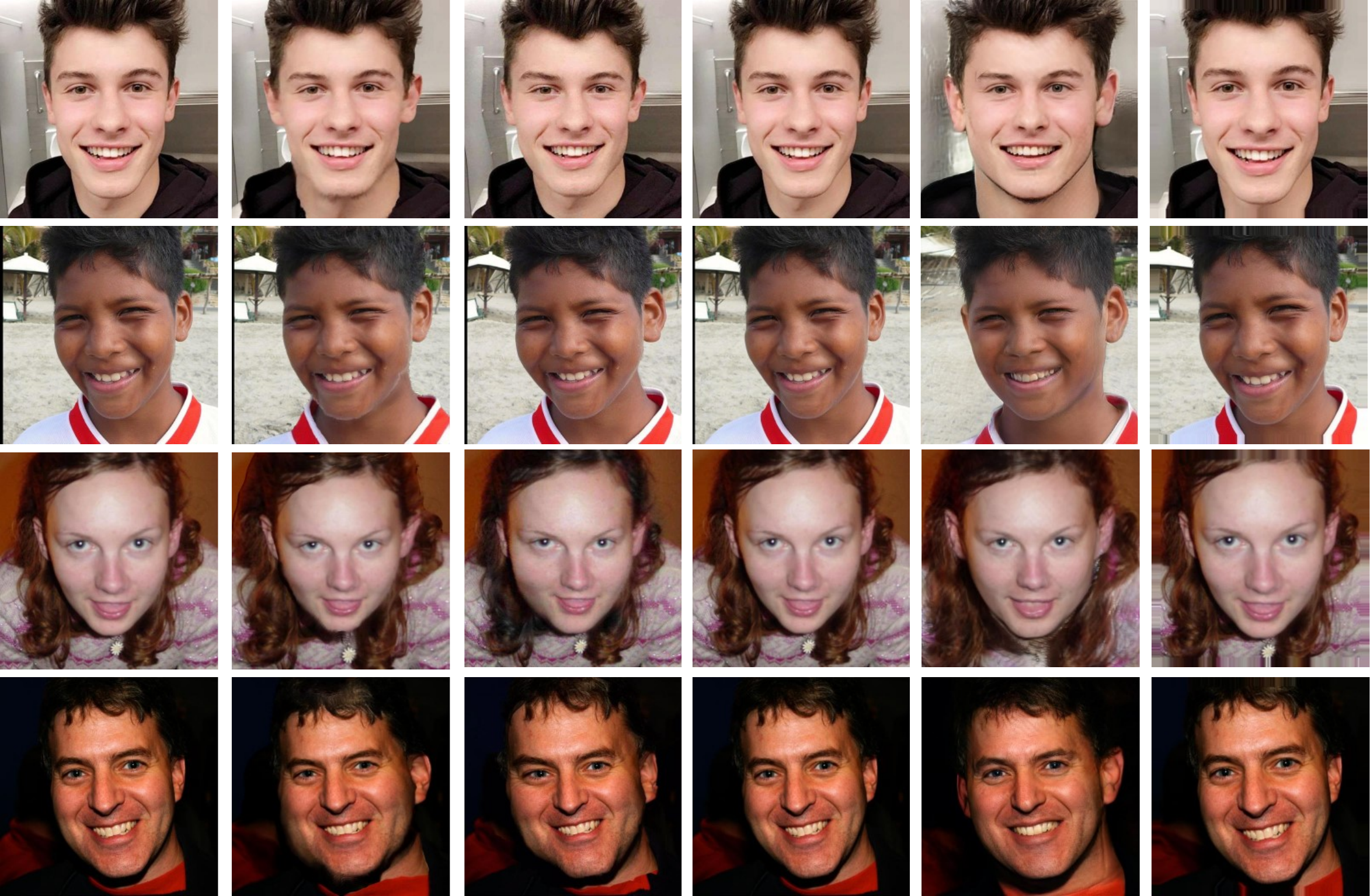}

\begin{tabularx}
{1\textwidth} { 
   >{\centering\arraybackslash}X 
   >{\centering\arraybackslash}X 
   >{\centering\arraybackslash}X 
   >{\centering\arraybackslash}X  
   >{\centering\arraybackslash}X 
   >{\centering\arraybackslash}X}
 Input & Ours & LPUP \cite{zhao2019learning} & Fried \cite{fried2016perspective} & Disco \cite{wang2023disco} & Shih \cite{shih2019distortion}  \\
\end{tabularx}

\caption{Qualitative comparison on in-the-wild images collected by LPUP \cite{zhao2019learning}.}

\label{fig:Comparison_wild_images_LPUP}
\end{figure*}

Next, we compare our methods qualitatively on in-the-wild images collected by LPUP \cite{zhao2019learning} as shown in Figure \ref{fig:Comparison_wild_images_LPUP}. Our method shows comparable results with the Disco method \cite{wang2023disco} while being far computationally less expensive. Moreover, in some cases, Disco's correction is exaggerated as in the first image. In other cases, it loses some details due to the inversion process while our method successfully retains the face and the other subject's details. For instance, this can be observed in the second image where the teeth of the child are visibly altered. Also, in the third image where the necklace of the subject is lost, the cloth details are wiped and the mouth-tongue shape is distorted. In comparison to LPUP \cite{zhao2019learning} and Fried \cite{fried2016perspective}, our method has a more faithful geometric rectification of the face as can be observed, for instance, in the chin of the subjects where more occluded regions from the neck appear when the camera moves far from the subject.

\begin{figure}

 \includegraphics[width=0.4\textwidth]{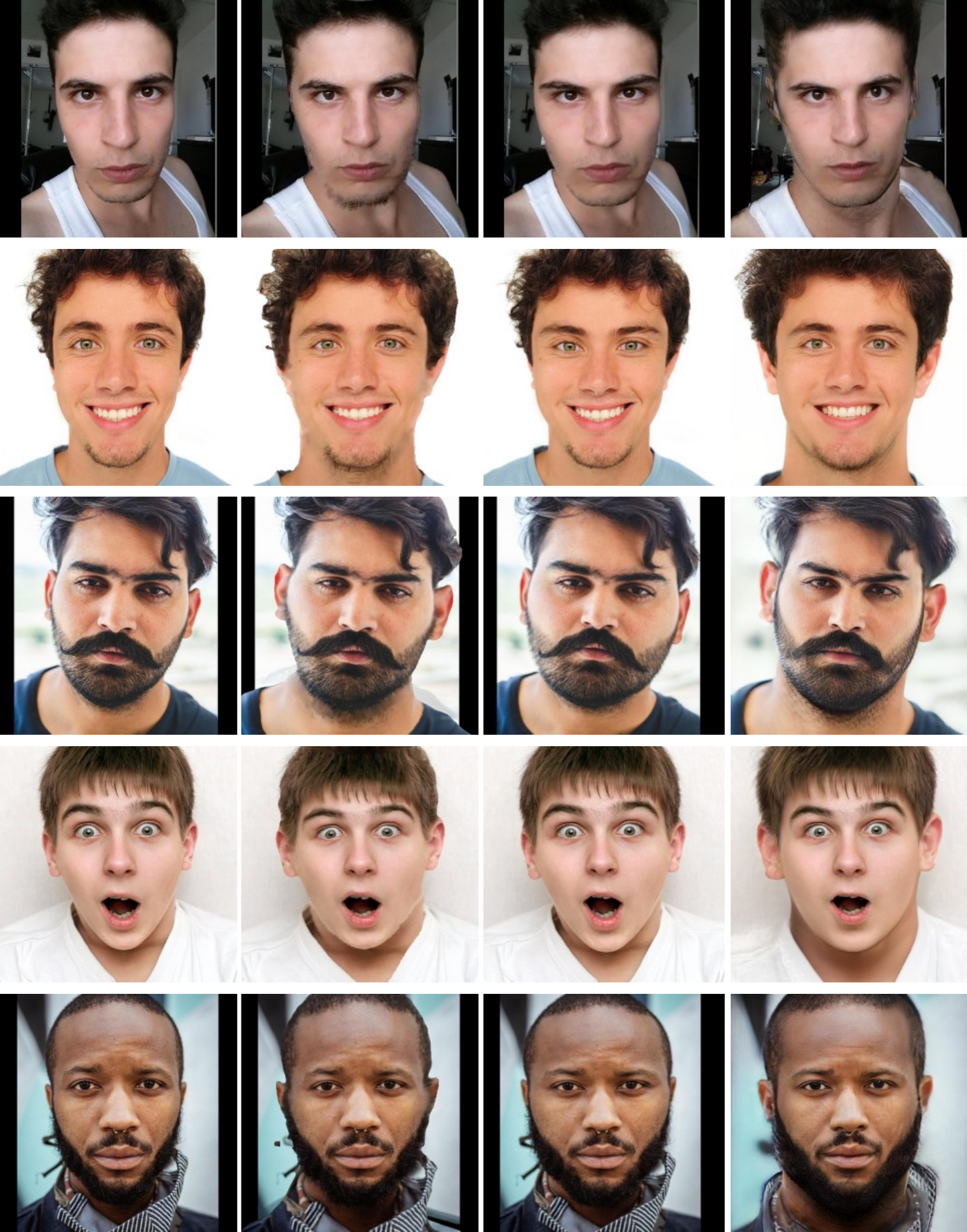}

\begin{tabularx}
{0.4\textwidth} { 
   >{\centering\arraybackslash}X 
   >{\centering\arraybackslash}X 
   >{\centering\arraybackslash}X 
   >{\centering\arraybackslash}X }
 Input & Ours & Fried \cite{fried2016perspective}  &  Disco \cite{wang2023disco} \\

\end{tabularx}
 % \framebox(\linewidth,300){}

\caption{Qualitative comparison on images collected by Disco \cite{wang2023disco}.}

\label{fig:Comparison_Disco_wild}
\end{figure}

Furthermore, Figure \ref{fig:Comparison_Disco_wild} shows some correction result comparisons on in-the-wild images collected by Disco \cite{wang2023disco}. Again, our method's rectification looks more faithful than Fried \cite{fried2016perspective}, and is quite comparable with Disco. It is important to note that we don't know the focal length of the cameras (or the field of view) in these cases, but it is likely to be quite different from the case used in generating the training data. Despite this, the method works well and clearly generalizes not only to real images but also to other configurations.

\begin{figure*}

\centering
 \includegraphics[width=.8\textwidth]{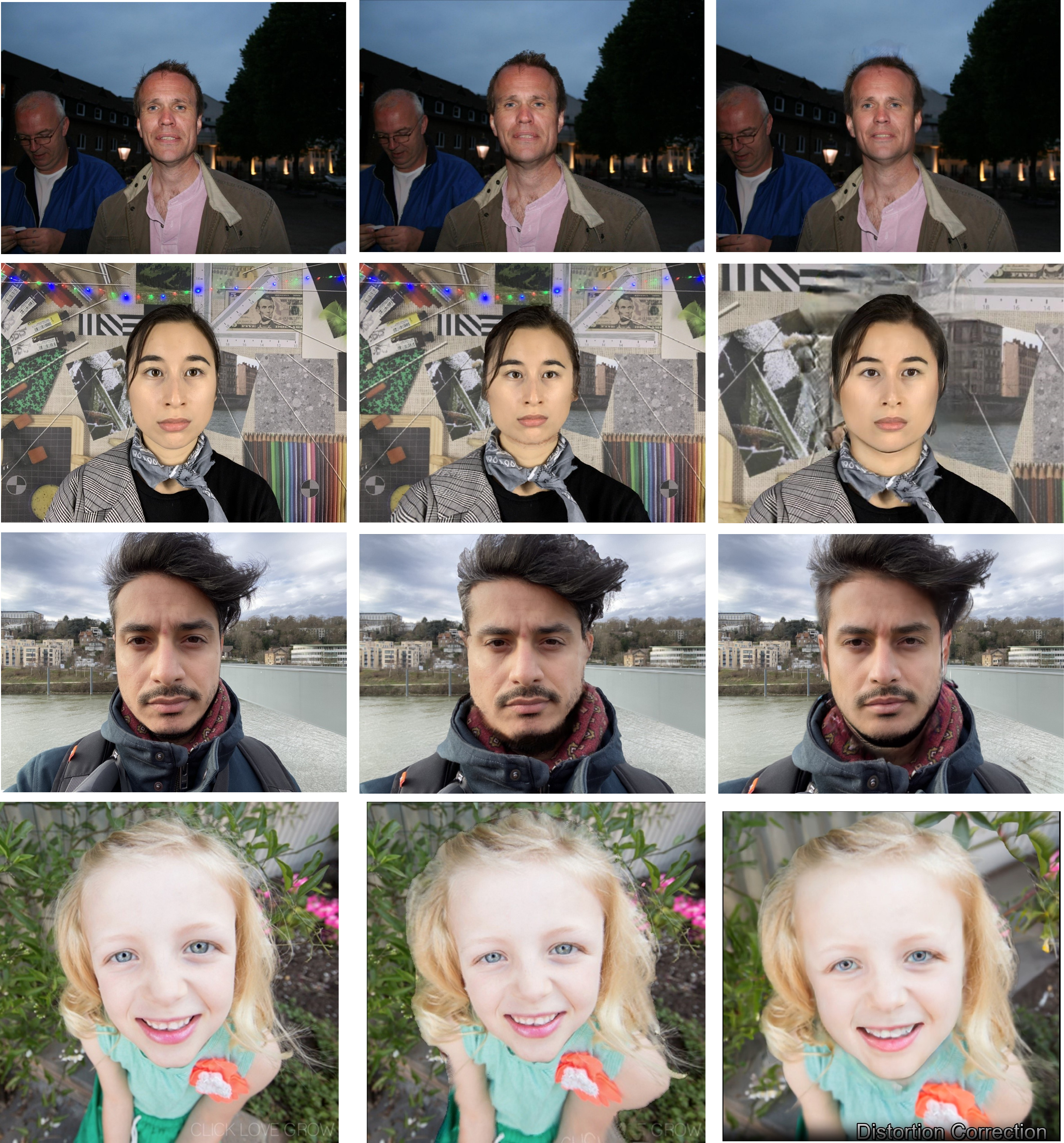}

\begin{tabularx}
{0.75\textwidth} { 
   >{\centering\arraybackslash}X 
   >{\centering\arraybackslash}X  
   >{\centering\arraybackslash}X }
 Input & Ours  & Disco \cite{wang2023disco} \\

\end{tabularx}

\caption{Full-frame image rectification comparison. The last image is a screenshot from the result website provided by the Disco method \cite{wang2023disco}.}

\label{fig:full_frame_comparison}
\end{figure*}

Finally, we compare the full-frame rectification quality with the Disco method \cite{wang2023disco}. Figure \ref{fig:full_frame_comparison} shows a qualitative comparison of full-frame results between our pipeline and  Disco \cite{wang2023disco}. Our method produces similar final results as Disco without separate processing of face crops.

\subsection{Ablation studies} 
\begin{figure}
\centering
 \includegraphics[width=0.5\textwidth]{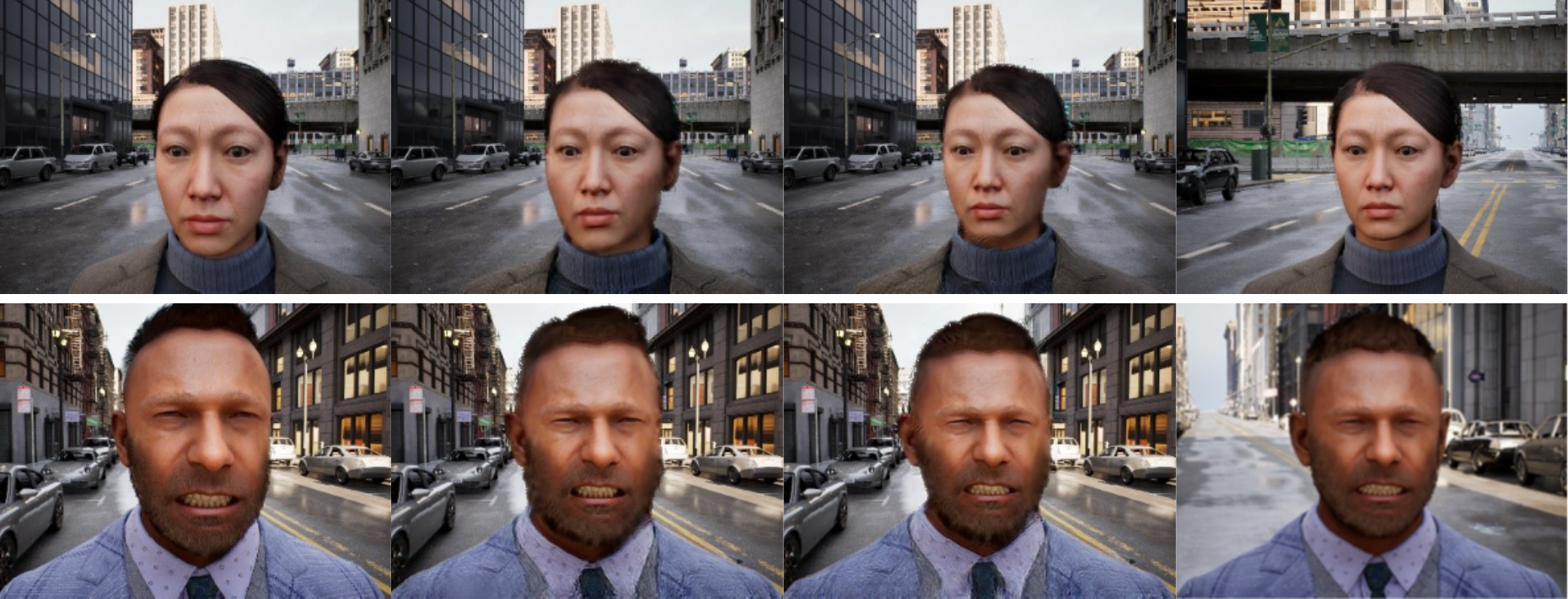}
 \begin{tabularx}
{0.5\textwidth} { 
   >{\centering\arraybackslash}X 
   >{\centering\arraybackslash}X  
   >{\centering\arraybackslash}X
   >{\centering\arraybackslash}X}
 Input & w/o E2E & Ours & GT \\

\end{tabularx}
\caption{Qualitative comparison on the synthetic dataset when disabling the E2E training of our depth estimation module.}

\label{fig:Ablating_E2E_synthetic}
\end{figure}

\noindent To show the effectiveness of our design choices. We ablate some components from our pipeline and compare the results to the full pipeline. First, we disable the end-to-end training of the depth estimation module and train it separately without back-propagating photometric losses from the generator module through the differentiable renderer. Table \ref{tab:ablating_e2e_HzT} shows the effect of training the depth network in an E2E fashion. The E2E training boosts the quality of the novel synthesized view of the subject in terms of PSNR and SSIM values. Furthermore, Figure \ref{fig:Ablating_E2E_synthetic}, shows the influence of the E2E training on the synthesized novel view on examples from the synthetic dataset. 

\begin{figure}
\centering
 \includegraphics[width=1.0\linewidth]{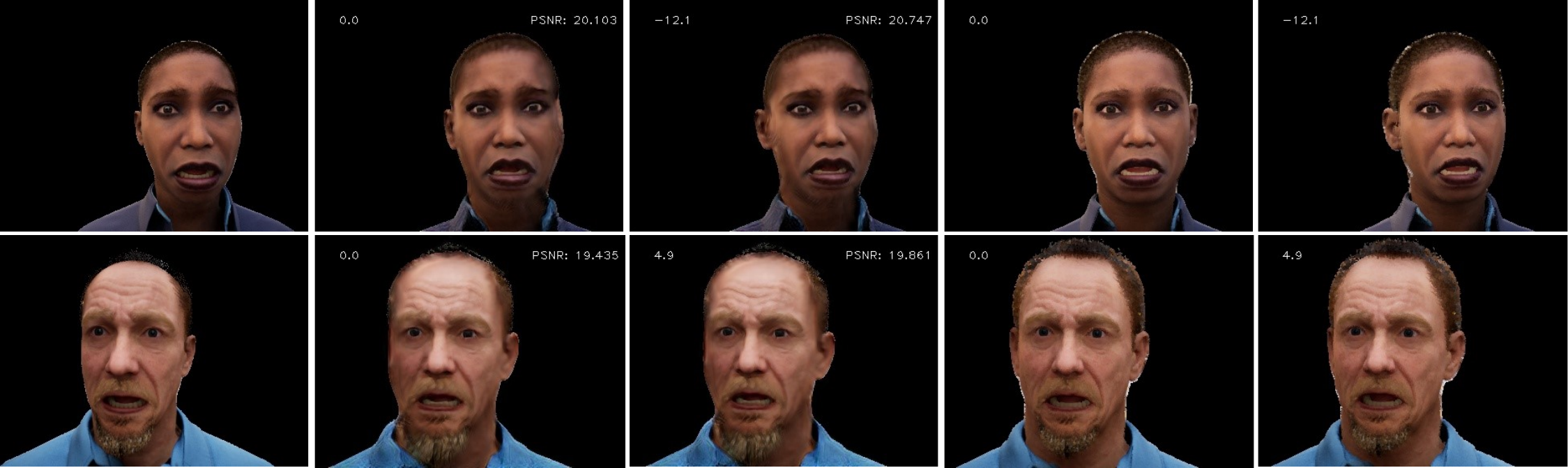}
 \begin{tabularx}{1.0\linewidth} { 
   >{\centering\arraybackslash}X 
   >{\centering\arraybackslash}X  
   >{\centering\arraybackslash}X
   >{\centering\arraybackslash}X
   >{\centering\arraybackslash}X}
 Input & w/o HzT & w/ HzT & GT $(t_x=0)$ & GT $(t_x\neq0)$ \\ 
\end{tabularx}

\caption{Qualitative comparison on the synthetic dataset when disabling the horizontal translation module (HzT). The predicted output and GT target are shown in both cases when $t_x= 0$ and the $t_x$ predicted by our model. The PSNR is reported on the right side of the image while the predicted $t_x$ is shown on the left.}

\label{fig:Ablating_HzT_Synthetic}
\end{figure}

\begin{table}
    \centering
\caption{Quantitative results on the synthetic multiview dataset for ablating the E2E training and the auxiliary translation module HzT.}

\label{tab:ablating_e2e_HzT}
\begin{tabularx}{1.\linewidth} { 
  l
  | >{\centering\arraybackslash}X
   >{\centering\arraybackslash}X
  | >{\centering\arraybackslash}X
  | >{\centering\arraybackslash}X  }
         \hline 
         Method&  E2E &HzT& PSNR $\uparrow$  &SSIM $\uparrow$  \\ \hline
         baseline &  & &18.813  & 0.7997      \\
         Ours w E2E &\checkmark  & & 19.617  & 0.8324     \\
         Ours full & \checkmark &\checkmark& 19.894  & 0.8332  \\ 
 \hline
\end{tabularx}
\end{table}

\begin{figure}
\centering
 \includegraphics[width=1.0\linewidth]{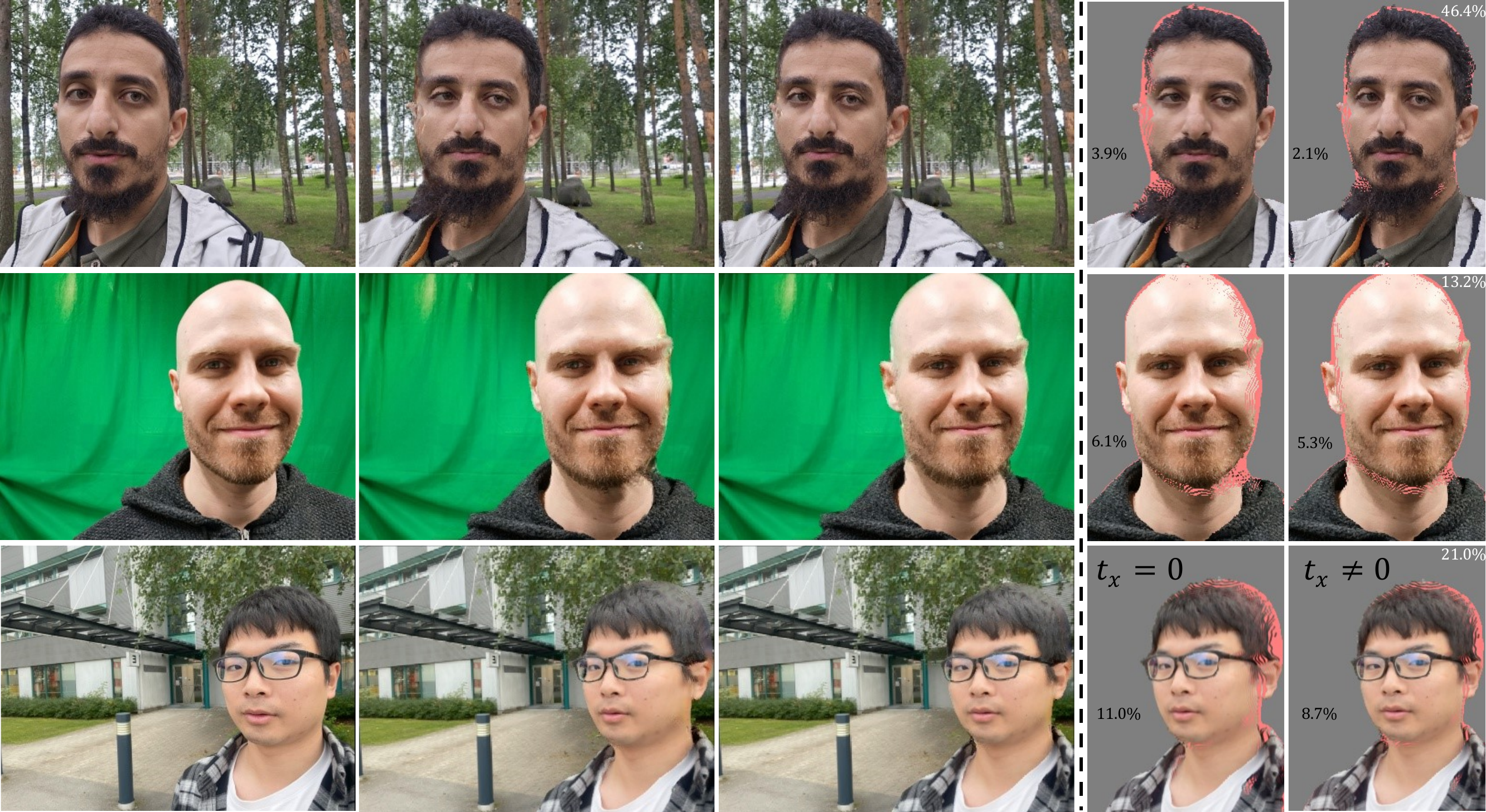}
 \begin{tabularx}
{1.0\linewidth} { 
   >{\centering\arraybackslash}X 
   >{\centering\arraybackslash}X  
   >{\centering\arraybackslash}X
   >{\centering\arraybackslash}X}
 Input & $t_x=0$ & $t_x\neq0$ & Inpainted region \\

\end{tabularx}
\caption{Qualitative comparison on the real images illustrating the effect of ablating the horizontal translation module. The orientation when $t_x\neq0$ gives the sense that there are no missing face parts such as ears. Additionally, the head orientation conforms with its corresponding input. The last two columns visualize the inpainted regions, highlighted in red, for each case. The percentage of the inpainted area is indicated on the left, while the reduction in inpainted area (when using horizontal translation) is shown in the top-right corner of each image.}

% The percentage of the inpainted area is shown on the left side, while the reduction percentage appears on the top right corner of the image.}

\label{fig:ablating_trans_real}
\end{figure}

Next, we demonstrate the benefit of incorporating the auxiliary translation module to enhance the synthesized novel views. Figure \ref{fig:Ablating_HzT_Synthetic} shows a comparison on the synthetic dataset when we ablate the translation module from our pipeline. The module can effectively predict a translation that avoids the hard missing regions in the face boosting the quality of the synthesized novel views in those cases. In Figure \ref{fig:ablating_trans_real}, we show the results of ablating the horizontal translation module on real data. Translating the camera horizontally gives the sense that there are no missing parts of the face such as the ear. In addition, the head orientation after translating the camera conforms with the orientation in its corresponding input. We also visualize the inpainted area for each case in the last two columns of Figure \ref{fig:ablating_trans_real}. It is shown that the inpainted region in the case of $t_x \neq 0$ is smaller. To demonstrate the generalizability of our modules trained on a synthetic dataset, we show images captured by different mobile phone cameras that have different focal lengths. The images of Figure \ref{fig:ablating_trans_real} are captured by Samsung S23 Ultra, Huawei P40Pro, and iPhone 12, respectively. Their 35mm-equivalent focal length are: 25, 26, and 23,  and their field of view angles are:  81.6°, 79.5°, and 86.5°. Our pipeline can deal with subjects wearing eyeglasses as shown in the third row of Figure \ref{fig:ablating_trans_real}.

Table \ref{tab:ablating_e2e_HzT} shows the quantitative results of disabling the translation module on the synthetic multi-view dataset.

\section{Limitations and Future Work}
\noindent It is observed in some cases that avoiding the hard missing areas of the face by the translation module is difficult. For example, the case when the face is in the centre of the image and both ears are occluded as in the image of the third row in Figure \ref{fig:full_frame_comparison}. Although Disco \cite{wang2023disco} and our methods hallucinate the missing ears quite reasonably, it still fails to faithfully inpaint them as shown in the left ear of the third example in Figure \ref{fig:full_frame_comparison}, and the first and last examples of Figure \ref{fig:Comparison_Disco_wild}. Utilizing a pretrained SOTA inpainting method is not usually successful for two reasons. First, we do not know the exact missing face region in advance. Second, the pretrained SOTA networks are influenced by the distribution of mask shapes during training. Hence, even if we provide a rough mask for the missing region for a pretrained network, the ears are not guaranteed to be inpainted in addition to the artefacts introduced on the face boundaries.

\section{Conclusion}
\noindent We proposed an E2E pipeline to rectify the perspective distortion in close-range portraits and selfie images. Facial depth prediction was predicted to undistort faces by adjusting the camera-to-subject distance and the focal length. We showed that modifying the horizontal translation can enhance the quality of reprojected images by decreasing the area that needs to be filled. Our feature extractor, depth predictor and generator nets were trained in an E2E fashion by leveraging a differentiable projection module. Our qualitative and quantitative results manifested the generalizability of our model on real images while being only trained with a synthetic dataset. This showed the effectiveness of the generated data in simulating real-world perspective distortion.  Finally, we showed that our method outperforms the previous methods of Fried \cite{fried2016perspective} and LPUP \cite{zhao2019learning}, and produces comparable results with Disco \cite{wang2023disco} while being 260 times faster.

\bibliographystyle{IEEEtran}
\bibliography{Bib}

\begin{thebibliography}{10}
\providecommand{\url}[1]{#1}
\csname url@samestyle\endcsname
\providecommand{\newblock}{\relax}
\providecommand{\bibinfo}[2]{#2}
\providecommand{\BIBentrySTDinterwordspacing}{\spaceskip=0pt\relax}
\providecommand{\BIBentryALTinterwordstretchfactor}{4}
\providecommand{\BIBentryALTinterwordspacing}{\spaceskip=\fontdimen2\font plus
\BIBentryALTinterwordstretchfactor\fontdimen3\font minus \fontdimen4\font\relax}
\providecommand{\BIBforeignlanguage}[2]{{%
\expandafter\ifx\csname l@#1\endcsname\relax
\typeout{** WARNING: IEEEtran.bst: No hyphenation pattern has been}%
\typeout{** loaded for the language `#1'. Using the pattern for}%
\typeout{** the default language instead.}%
\else
\language=\csname l@#1\endcsname
\fi
#2}}
\providecommand{\BIBdecl}{\relax}
\BIBdecl

\bibitem{zhao2019learning}
Y.~Zhao, Z.~Huang, T.~Li, W.~Chen, C.~LeGendre, X.~Ren, A.~Shapiro, and H.~Li, ``Learning perspective undistortion of portraits,'' in \emph{Proceedings of the IEEE/CVF International Conference on Computer Vision}, 2019, pp. 7849--7859.

\bibitem{kao2023toward}
Y.~Kao, B.~Pan, M.~Xu, J.~Lyu, X.~Zhu, Y.~Chang, X.~Li, and Z.~Lei, ``Toward 3d face reconstruction in perspective projection: Estimating 6dof face pose from monocular image,'' \emph{IEEE Transactions on Image Processing}, vol.~32, pp. 3080--3091, 2023.

\bibitem{lai2021correcting}
W.-S. Lai, Y.~Shih, C.-K. Liang, and M.-H. Yang, ``Correcting face distortion in wide-angle videos,'' \emph{IEEE Transactions on Image Processing}, vol.~31, pp. 366--378, 2021.

\bibitem{shih2019distortion}
Y.~Shih, W.-S. Lai, and C.-K. Liang, ``Distortion-free wide-angle portraits on camera phones,'' \emph{ACM Transactions on Graphics (TOG)}, vol.~38, no.~4, pp. 1--12, 2019.

\bibitem{ravi2020pytorch3d}
N.~Ravi, J.~Reizenstein, D.~Novotny, T.~Gordon, W.-Y. Lo, J.~Johnson, and G.~Gkioxari, ``Accelerating 3d deep learning with pytorch3d,'' \emph{arXiv:2007.08501}, 2020.

\bibitem{fried2016perspective}
O.~Fried, E.~Shechtman, D.~B. Goldman, and A.~Finkelstein, ``Perspective-aware manipulation of portrait photos,'' \emph{ACM Transactions on Graphics (TOG)}, vol.~35, no.~4, pp. 1--10, 2016.

\bibitem{wang2023disco}
Z.~Wang, Y.-L. Liu, J.-B. Huang, S.~Satoh, S.~Ma, G.~Krishnan, and J.~Wang, ``Disco: Portrait distortion correction with perspective-aware 3d gans,'' \emph{International Journal of Computer Vision}, vol. 132, no.~11, pp. 5471--5488, 2024.

\bibitem{yin2021learning}
W.~Yin, J.~Zhang, O.~Wang, S.~Niklaus, L.~Mai, S.~Chen, and C.~Shen, ``Learning to recover 3d scene shape from a single image,'' in \emph{Proceedings of the IEEE/CVF Conference on Computer Vision and Pattern Recognition}, 2021, pp. 204--213.

\bibitem{zhang2013high}
X.~Zhang, L.~Yin, J.~F. Cohn, S.~Canavan, M.~Reale, A.~Horowitz, and P.~Liu, ``A high-resolution spontaneous 3d dynamic facial expression database,'' in \emph{2013 10th IEEE international conference and workshops on automatic face and gesture recognition (FG)}.\hskip 1em plus 0.5em minus 0.4em\relax IEEE, 2013, pp. 1--6.

\bibitem{unrealengine}
\BIBentryALTinterwordspacing
{Epic Games}, ``{Unreal Engine},'' 2022. [Online]. Available: \url{https://www.unrealengine.com}
\BIBentrySTDinterwordspacing

\bibitem{luo2016hole}
G.~Luo, Y.~Zhu, Z.~Li, and L.~Zhang, ``A hole filling approach based on background reconstruction for view synthesis in 3d video,'' in \emph{Proceedings of the IEEE conference on computer vision and pattern recognition}, 2016, pp. 1781--1789.

\bibitem{shih20203d}
M.-L. Shih, S.-Y. Su, J.~Kopf, and J.-B. Huang, ``3d photography using context-aware layered depth inpainting,'' in \emph{Proceedings of the IEEE/CVF Conference on Computer Vision and Pattern Recognition}, 2020, pp. 8028--8038.

\bibitem{wiles2020synsin}
O.~Wiles, G.~Gkioxari, R.~Szeliski, and J.~Johnson, ``Synsin: End-to-end view synthesis from a single image,'' in \emph{Proceedings of the IEEE/CVF Conference on Computer Vision and Pattern Recognition}, 2020, pp. 7467--7477.

\bibitem{nagano2019deep}
K.~Nagano, H.~Luo, Z.~Wang, J.~Seo, J.~Xing, L.~Hu, L.~Wei, and H.~Li, ``Deep face normalization,'' \emph{ACM Transactions on Graphics (TOG)}, vol.~38, no.~6, pp. 1--16, 2019.

\bibitem{chan2022efficient}
E.~R. Chan, C.~Z. Lin, M.~A. Chan, K.~Nagano, B.~Pan, S.~De~Mello, O.~Gallo, L.~J. Guibas, J.~Tremblay, S.~Khamis \emph{et~al.}, ``Efficient geometry-aware 3d generative adversarial networks,'' in \emph{Proceedings of the IEEE/CVF conference on computer vision and pattern recognition}, 2022, pp. 16\,123--16\,133.

\bibitem{yang20233dhumangan}
Z.~Yang, S.~Li, W.~Wu, and B.~Dai, ``3dhumangan: 3d-aware human image generation with 3d pose mapping,'' in \emph{Proceedings of the IEEE/CVF International Conference on Computer Vision}, 2023, pp. 23\,008--23\,019.

\bibitem{roich2022pivotal}
D.~Roich, R.~Mokady, A.~H. Bermano, and D.~Cohen-Or, ``Pivotal tuning for latent-based editing of real images,'' \emph{ACM Transactions on graphics (TOG)}, vol.~42, no.~1, pp. 1--13, 2022.

\bibitem{zhang2018perceptual}
R.~Zhang, P.~Isola, A.~A. Efros, E.~Shechtman, and O.~Wang, ``The unreasonable effectiveness of deep features as a perceptual metric,'' in \emph{CVPR}, 2018.

\bibitem{karpikova2024super}
P.~Karpikova, A.~Spiridonov, A.~Vorontsova, A.~Yaschenko, E.~Radionova, I.~Medvedev, and A.~Limonov, ``Super: Selfie undistortion and head pose editing with identity preservation,'' in \emph{2024 IEEE International Conference on Image Processing (ICIP)}.\hskip 1em plus 0.5em minus 0.4em\relax IEEE, 2024, pp. 1704--1710.

\bibitem{Chen_2024_CVPR}
B.~Chen, B.~Curless, I.~Kemelmacher-Shlizerman, and S.~M. Seitz, ``Total selfie: Generating full-body selfies,'' in \emph{Proceedings of the IEEE/CVF Conference on Computer Vision and Pattern Recognition (CVPR)}, June 2024, pp. 6701--6711.

\bibitem{wang2021one}
T.-C. Wang, A.~Mallya, and M.-Y. Liu, ``One-shot free-view neural talking-head synthesis for video conferencing,'' in \emph{Proceedings of the IEEE/CVF conference on computer vision and pattern recognition}, 2021, pp. 10\,039--10\,049.

\bibitem{gao2020portrait}
C.~Gao, Y.~Shih, W.-S. Lai, C.-K. Liang, and J.-B. Huang, ``Portrait neural radiance fields from a single image,'' \emph{arXiv preprint arXiv:2012.05903}, 2020.

\bibitem{chen2022robust}
X.~Chen, Y.~Zhu, Y.~Li, B.~Fu, L.~Sun, Y.~Shan, and S.~Liu, ``Robust human matting via semantic guidance,'' in \emph{Proceedings of the Asian Conference on Computer Vision}, 2022, pp. 2984--2999.

\bibitem{he2016deep}
K.~He, X.~Zhang, S.~Ren, and J.~Sun, ``Deep residual learning for image recognition,'' in \emph{Proceedings of the IEEE conference on computer vision and pattern recognition}, 2016, pp. 770--778.

\bibitem{adelson1984pyramid}
E.~H. Adelson, C.~H. Anderson, J.~R. Bergen, P.~J. Burt, and J.~M. Ogden, ``Pyramid methods in image processing,'' \emph{RCA engineer}, vol.~29, no.~6, pp. 33--41, 1984.

\bibitem{li2022mat}
W.~Li, Z.~Lin, K.~Zhou, L.~Qi, Y.~Wang, and J.~Jia, ``Mat: Mask-aware transformer for large hole image inpainting,'' in \emph{Proceedings of the IEEE/CVF conference on computer vision and pattern recognition}, 2022, pp. 10\,758--10\,768.

\bibitem{johnson2016perceptual}
J.~Johnson, A.~Alahi, and L.~Fei-Fei, ``Perceptual losses for real-time style transfer and super-resolution,'' in \emph{Computer Vision--ECCV 2016: 14th European Conference, Amsterdam, The Netherlands, October 11-14, 2016, Proceedings, Part II 14}.\hskip 1em plus 0.5em minus 0.4em\relax Springer, 2016, pp. 694--711.

\bibitem{wang2018high}
T.-C. Wang, M.-Y. Liu, J.-Y. Zhu, A.~Tao, J.~Kautz, and B.~Catanzaro, ``High-resolution image synthesis and semantic manipulation with conditional gans,'' in \emph{Proceedings of the IEEE conference on computer vision and pattern recognition}, 2018, pp. 8798--8807.

\bibitem{kingma2014adam}
D.~P. Kingma and J.~Ba, ``Adam: A method for stochastic optimization,'' \emph{arXiv preprint arXiv:1412.6980}, 2014.

\bibitem{wang2004image}
Z.~Wang, A.~C. Bovik, H.~R. Sheikh, and E.~P. Simoncelli, ``Image quality assessment: from error visibility to structural similarity,'' \emph{IEEE transactions on image processing}, vol.~13, no.~4, pp. 600--612, 2004.

\bibitem{lugaresi2019mediapipe}
C.~Lugaresi, J.~Tang, H.~Nash, C.~McClanahan, E.~Uboweja, M.~Hays, F.~Zhang, C.-L. Chang, M.~Yong, J.~Lee \emph{et~al.}, ``Mediapipe: A framework for perceiving and processing reality,'' in \emph{Third workshop on computer vision for AR/VR at IEEE computer vision and pattern recognition (CVPR)}, vol. 2019, 2019.

\bibitem{yao2024combining}
L.~Yao, C.~Chen, X.~Li, Z.~Yan, and W.~Zuo, ``Combining generative and geometry priors for wide-angle portrait correction,'' in \emph{European Conference on Computer Vision}.\hskip 1em plus 0.5em minus 0.4em\relax Springer, 2024, pp. 395--411.

\bibitem{burgos2014distance}
X.~P. Burgos-Artizzu, M.~R. Ronchi, and P.~Perona, ``Distance estimation of an unknown person from a portrait,'' in \emph{Computer Vision--ECCV 2014: 13th European Conference, Zurich, Switzerland, September 6-12, 2014, Proceedings, Part I 13}.\hskip 1em plus 0.5em minus 0.4em\relax Springer, 2014, pp. 313--327.

\end{thebibliography}

\begin{IEEEbiography}[{\includegraphics[width=1in,height=1.25in,clip,keepaspectratio]{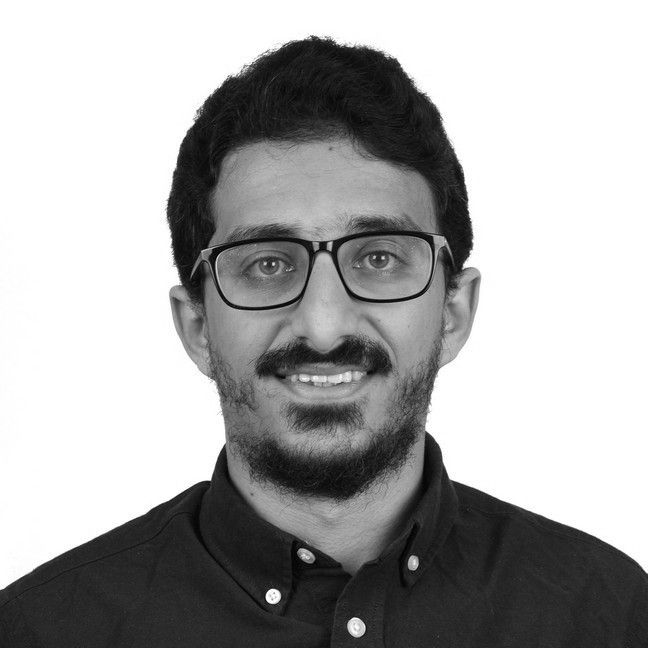}}]{Ahmed Alhawwary} received the BSs degree in computer and systems engineering from Faculty of Engineering, Alexandria University, Egypt and the MSc degree in computer science and engineering from the Faculty of Information Technology and Electrical Engineering, the University of Oulu, Finland. Currently, he is working as a doctoral candidate with the Center for Machine Vision and Signal Analysis. His research interests include computer vision and deep learning. 
\end{IEEEbiography}

\begin{IEEEbiography}[{\includegraphics[width=1in,height=1.25in,clip,keepaspectratio]{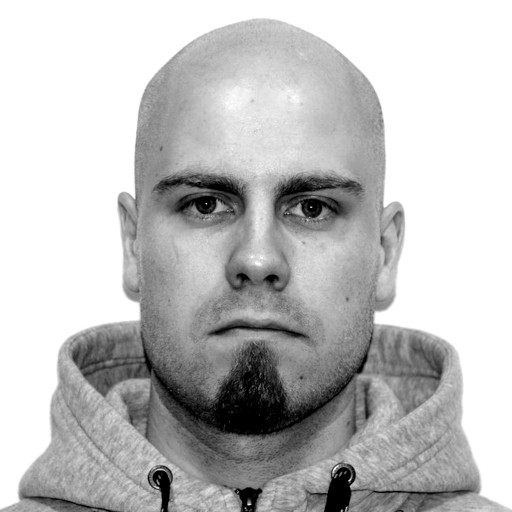}}]{Janne Mustaniemi} received the MSc degree in computer science and engineering from the University of Oulu, Finland. Later in 2020, he received the Doctor of Science degree in computer science and engineering from the same university. He is currently working as a postdoctoral researcher in the Center for Machine Vision and Signal Analysis (CMVS) at the University of Oulu. His research interests include 3D computer vision, computational photography, and machine learning.
\end{IEEEbiography}

\begin{IEEEbiography}[{\includegraphics[width=1in,height=1.25in,clip,keepaspectratio]{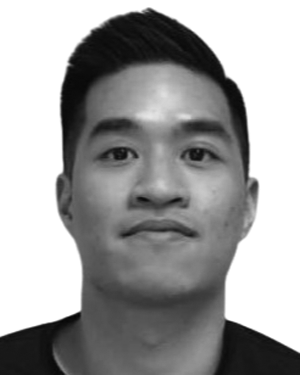}}]{Phong Nguyen-Ha}
received the BSs degree in mechanical engineering from the Ha Noi University of Science and Technology (HUST), Vietnam and the MSc degree in computer science engineering from Dongguk University, Seoul, South Korea. Since then, he is working as a doctoral candidate with CMVS, fully funded by the Vision-based 3D perception for mixed reality applications grant from the Infotech Institute. His research interests include 3D computer vision, computer graphics and deep learning.
\end{IEEEbiography}

\begin{IEEEbiography}[{\includegraphics[width=1in,height=1.25in,clip,keepaspectratio]{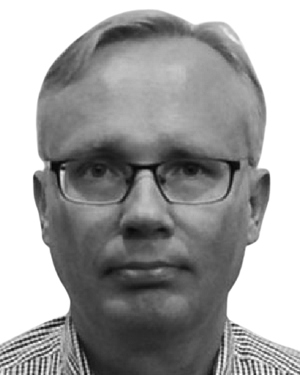}}]{Janne Heikkilä}
received the Doctor of Science
in Technology degree in information engineering
from the University of Oulu, Oulu, Finland, in
1998. He is currently a Professor of computer
vision and digital video processing with the Faculty of Information Technology and Electrical Engineering, University of Oulu, and the Head of
the Degree Program in computer science and
engineering. He has supervised nine completed
doctoral dissertations and authored/coauthored
more than 160 peer-reviewed scientific articles
in international journals and conferences. His research interests include
computer vision, machine learning, digital image and video processing,
and biomedical image analysis. Prof. Heikkilä has served as an Area
Chair and a member of program and organizing committees of several
international conferences. He is a Senior Editor for the Journal of
Electronic Imaging, an Associate Editor for the IET Computer Vision
and Electronic Letters on Computer Vision and Image Processing, a
Guest Editor for a special issue in Multimedia Tools and Applications,
and a member of the Governing Board of the International Association
for Pattern Recognition. During 2006–2009, he was the President of
the Pattern Recognition Society of Finland. He has been the Principal
Investigator in numerous research projects funded by the Academy of
Finland and the National Agency for Technology and Innovation. \end{IEEEbiography}

\end{document}